\documentclass[12pt, onecolumn, journal]{IEEEtran}

\usepackage{amsmath,graphicx}
\usepackage{amsfonts}
\usepackage{cite}
\usepackage{bm}
\usepackage{caption2}
\usepackage[numbers,sort&compress]{natbib}
\usepackage{algorithmicx}
\usepackage[ruled]{algorithm}
\usepackage{algpseudocode}
\usepackage{multirow}
\usepackage{amssymb}
\usepackage{color}
\begin{document}

\renewcommand{\algorithmicrequire}{\textbf{Input:}}
\renewcommand{\algorithmicensure}{\textbf{Output:}}
\algnewcommand\algorithmicswitch{\textbf{switch}}
\algnewcommand\algorithmiccase{\textbf{case}}
\algnewcommand\algorithmicassert{\texttt{assert}}
\algnewcommand\Assert[1]{\State \algorithmicassert(#1)}%
% New "environments"
\algdef{SE}[SWITCH]{Switch}{EndSwitch}[1]{\algorithmicswitch\ #1\ \algorithmicdo}{\algorithmicend\ \algorithmicswitch}%
\algdef{SE}[CASE]{Case}{EndCase}[1]{\algorithmiccase\ #1}{\algorithmicend\ \algorithmiccase}%
\algtext*{EndSwitch}%
\algtext*{EndCase}%

\title{Compressive Deconvolution in Medical Ultrasound Imaging}

\author{\IEEEauthorblockN{Zhouye Chen, Adrian Basarab, Denis Kouam\'{e}}\\
\IEEEauthorblockA{University of Toulouse, IRIT UMR CNRS 5505, Toulouse, France}
\thanks{}}

\author{Zhouye~Chen,~\IEEEmembership{Student Member,~IEEE,}
        Adrian~Basarab,~\IEEEmembership{Member,~IEEE,}
        and~Denis~Kouam\'{e},~\IEEEmembership{Member,~IEEE}% <-this % stops a space
\thanks{Zhouye Chen, Adrian Basarab and Denis Kouam\'{e} are with IRIT UMR CNRS 5505, University of Toulouse, Toulouse, France (e-mail:{zhouye.chen,adrian.basarab,denis.kouame}@irit.fr).}}% <-this % stops a space

%\markboth{Accepted by IEEE TRANSACTIONS ON MEDICAL IMAGING}%
%{Chen \MakeLowercase{\textit{et al.}}: Compressive Deconvolution in Medical Ultrasound Imaging}

%\IEEEpubid{\begin{minipage}{\textwidth}\ \\\begin{center}
%  Copyright\copyright~2010 IEEE. Personal use of this material is permitted.\\ 
%  However, permission to use this material for any other purposes must be obtained from the IEEE by sending a request to pubs-permissions@ieee.org\end{center}
%\end{minipage}}

\IEEEtitleabstractindextext{%
\begin{abstract}
The interest of compressive sampling in ultrasound imaging has been recently extensively evaluated by several research teams. Following the different application setups, it has been shown that the RF data may be reconstructed from a small number of measurements and/or using a reduced number of ultrasound pulse emissions. Nevertheless, RF image spatial resolution, contrast and signal to noise ratio are affected by the limited bandwidth of the imaging transducer and the physical phenomenon related to US wave propagation. To overcome these limitations, several deconvolution-based image processing techniques have been proposed to enhance the ultrasound images. In this paper, we propose a novel framework, named compressive deconvolution, that reconstructs enhanced RF images from compressed measurements. Exploiting an unified formulation of the direct acquisition model, combining random projections and 2D convolution with a spatially invariant point spread function, the benefit of our approach is the joint data volume reduction and image quality improvement. The proposed optimization method, based on the Alternating Direction Method of Multipliers, is evaluated on both simulated and \textit{in vivo} data.
\end{abstract}

}

\maketitle

\IEEEdisplaynontitleabstractindextext

\IEEEpeerreviewmaketitle

\section{Introduction}
\label{sec1}
\IEEEPARstart{U}{ltrasound} (US) medical imaging has the advantages of being noninvasive, harmless, cost-effective and portable over other imaging modalities such as X-ray Computed Tomography or Magnetic Resonance Imaging \cite{szabo2004diagnostic}. 

Despite its intrinsic rapidity of acquisition, several US applications such as cardiac, Doppler, elastography or 3D imaging may require higher frame rates than those provided by conventional acquisition schemes (e.g. ultrafast imaging \cite{tanter2014ultrafast}) or may suffer from the high amount of acquired data. In this context, a few research teams have recently evaluated the application of compressive sampling (CS) to 2D and 3D US imaging (e.g. \cite{achim2010compressive,quinsac2012frequency,chernyakova2014fourier, liebgott2012compressive,liebgott2013pre,schiffner2014pulse}) or to duplex Doppler \cite{RICH-11}. CS is a mathematical framework allowing to recover, via non linear optimization routines, an image from few linear measurements (below the limit standardly imposed by the Shannon-Nyquist theorem) \cite{donoho2006compressed,candes2006robust}. The CS acquisition model is given by

\begin{equation}
\boldsymbol{y} = \Phi\boldsymbol{r} + \boldsymbol{n} \label{CSDefinition}
\end{equation}

where $\boldsymbol{y}\in \mathbb{R}^{M}$ corresponds to the $M$ compressed measurements of the image $\boldsymbol{r}\in \mathbb{R}^{N}$ (one US radiofrequency (RF) image in our case), $\Phi\in \mathbb{R}^{M\times N}$ represents the CS acquisition matrix composed for example of $M$ random Gaussian vectors with $M<<N$ and $\boldsymbol{n}\in \mathbb{R}^{M}$ stands for a zero-mean additive white Gaussian noise. 

\IEEEpubidadjcol

The CS theory demonstrates that the $N$ pixels of image $\boldsymbol{r}$ may be recovered from the $M$ measurements in $\boldsymbol{y}$ provided two conditions: i) the image must have a sparse representation in a known basis or frame and ii) the measurement matrix and spasifying basis must be incoherent \cite{candes2008introduction}. In US imaging, despite the difficulties of sparsifying the data because of the speckle noise, it has been shown that RF images may be recovered in basis such as 2D Fourier \cite{quinsac2012frequency}, wavelets \cite{schiffner2014pulse}, waveatoms \cite{liebgott2013pre} or learning dictionaries \cite{LORI-15}, considering Bernoulli Gaussian \cite{dobigeon2012regularized} or $\alpha$-stable statistical assumptions \cite{basarab2014medical} and using various acquisition schemes such as plane-wave \cite{schiffner2014pulse}, Xampling \cite{chernyakova2014fourier} or projections on Gaussian \cite{achim2010compressive} or Bernoulli random vectors \cite{quinsac2012frequency}.  

However, the existing methods of CS in US have been shown to be able to recover images with a quality at most equivalent to those acquired using standard schemes. Nevertheless, the spatial resolution, the signal-to-noise ratio and the contrast of standard US images ($\boldsymbol{r}$ in eq. \eqref{CSDefinition}) are affected by the limited bandwidth of the imaging transducer, the physical phenomenon related to US wave propagation such as the diffraction and the imaging system. In order to overcome these issues, one of the research tracks extensively explored in the literature is the deconvolution of US images \cite{taxt2001two,michailovich2007blind,morin2013semi, zhao2014restoration,alessandrini2011restoration,alessandrini2011statistical}. Based on the first order Born approximation, these methods assume that the US RF images follow a 2D convolution model between the point spread function (PSF) and the tissue reflectivity function (TRF), \textit{i.e.} the image to be recovered \cite{jensen1993deconvolution}. Specifically, this results in $\boldsymbol{r}=H\boldsymbol{x}$, where $H \in \mathbb{R}^{N\times N}$ is a block circulant with circulant block (BCCB) matrix related to the 2D PSF of the system and $\boldsymbol{x}\in \mathbb{R}^{N}$ represents the lexicographically ordered tissue reflectivity function. We emphasize that this convolution model is based on the assumption of spatially invariant PSF. Although the PSF of an ultrasound system varies spatially and mostly in the axial direction of the images, many settings of the ultrasound imaging system allow to attenuate this spatial variation in practice, such as the dynamic focalization of the received echoes, the multiple focusing in emission and the time gain compensation (TGC). For this reason, the PSF is considered spatially invariant in our work, as it is the case in part of the existing works on image deconvolution in ultrasound imaging (e.g. \cite{michailovich2007blind, alessandrini2011restoration}).

The objective of our paper is to propose a novel technique which is able to jointly achieve US data volume reduction and image quality improvement. In other words, the main idea is to combine the two frameworks of CS and deconvolution applied to US imaging, resulting in the so-called compressive deconvolution (or CS deblurring) problem \cite{ma2009deblurring, xiao2011compounded, zhao2010compressed, amizic2013compressive,spinoulas2012simultaneous}. The combined direct model of joint CS and deconvolution is as follows:

\begin{equation}
\boldsymbol{y} = \Phi H\boldsymbol{x} + \boldsymbol{n} \label{eq1}
\end{equation}

where the variables $\boldsymbol{y}$, $\Phi$, $H$, $\boldsymbol{x}$ and $\boldsymbol{n}$ have the same meaning as defined above. Inverting the model in \eqref{eq1} will allow us to estimate the TRF $\boldsymbol{x}$ from the compressed RF measurements $\boldsymbol{y}$. 

To our knowledge, our work is the first attempt of addressing the compressive deconvolution problem in US imaging. In the general-purpose image processing literature, a few methods have been already proposed aiming at solving \eqref{eq1} \cite{hegde2009compressive,5677611,zhao2010compressed,
ma2009deblurring, xiao2011compounded, amizic2013compressive,spinoulas2012simultaneous}. In \cite{hegde2009compressive,5677611,zhao2010compressed}, the authors assumed $\boldsymbol{x}$ was sparse in the direct or image domain and the PSF was unknown. In \cite{hegde2009compressive,5677611}, a study on the number of measurements lower bound is presented, together with an algorithm to estimate the PSF and $\boldsymbol{x}$ alternatively. The authors in \cite{zhao2010compressed} solved the compressive deconvolution problem using an $\ell_1$-norm minimization algorithm by making use of the "all-pole" model of the autoregressive process. In \cite{
ma2009deblurring, xiao2011compounded}, $\boldsymbol{x}$ was considered sparse in a transformed domain and the PSF was supposed known. An algorithm based on Poisson singular integral and iterative curvelet thresholding was shown in \cite{
ma2009deblurring}. The authors in \cite{xiao2011compounded} further combined the curvelet regularization with total variation to improve the performance in \cite{
ma2009deblurring}. Finally, the methods in \cite{amizic2013compressive,spinoulas2012simultaneous} supposed the blurred signal $\boldsymbol{r}=H\boldsymbol{x}$ was sparse in a transformed domain and the PSF unknown. They proposed a compressive deconvolution framework that relies on a constrained optimization technique allowing to exploit existing CS reconstruction algorithms.

In this paper, we propose a compressive deconvolution technique adapted to US imaging. Our solution is based on the alternating direction method of multipliers (ADMM) \cite{boyd2011distributed, zhao2015alternating} and exploits two constraints. The first one, inspired from CS, imposes via an $l_1$-norm the sparsity of the RF image $\boldsymbol{r}$ in the transformed domain (Fourier domain, wavelet domain, etc). The second one imposes \textit{a priori} information for the TRF $\boldsymbol{x}$. Gaussian and Laplacian statistics have been extensively explored in US imaging (see e.g. \cite{michailovich2007blind, jirik2008blind, yu2012blind}. Moreover, recent results show that the Generalized Gaussian Distributed (GGD) is well adapted to model the TRF. Consequently, we employ herein the minimization of an $l_p$-norm of $\boldsymbol{x}$, covering all possible cases ranging from 1 to 2 \cite{alessandrini2011restoration, zhao2014restoration,7174535}. Similar to all existing frameworks, we consider the CS sampling matrix $\Phi$ known. In US imaging, the PSF is unknown in practical applications. However, its estimation from the RF data as an initialization step for the deconvolution has been extensively explored in US imaging. In this paper, we adopted the approach in \cite{michailovich2005novel} in order to estimate the PSF further used to construct the matrix $H$.

This paper is organized as follows. In section \ref{sec2}, we formulate our problem as a convex optimization routine and propose an ADMM-based method to efficiently solve it. Supporting simulated and experimental results are provided in section \ref{sec3} showing the contribution of our approach compared to existing methods and its efficiency in recovering the TRF from compressed US data. The conclusions are drawn in \ref{sec4}.

\section{Proposed Ultrasound Compressive Deconvolution Algorithm}
\label{sec2}
\subsection{Optimization Problem Formulation}
\subsubsection{Sequential approach}
In order to estimate the TRF $\boldsymbol{x}$ from the compressed and blurred measurements $\boldsymbol{y}$, an intuitive idea to invert the direct model in (\ref{eq1}) is to proceed through two sequential steps. The aim of the first step is to recover the blurred US RF image $\boldsymbol{r}=H\boldsymbol{x}$
from the compressed measurements $\boldsymbol{y}$ by solving the following optimization problem:

\begin{equation}
\underset{\boldsymbol{a}\in \mathbb{R}^{N}}{min}\left \| \boldsymbol{a} \right \|{_{1}}+\frac{1}{2\mu}\left \|\boldsymbol{y}-\Phi \Psi \boldsymbol{a}  \right \|{_{2}^{2}} \label{eq2}
\end{equation}

where $\boldsymbol{a}$ is the sparse representation of the US RF image $\boldsymbol{r}$ in the transformed domain $\Psi$, that is, $\boldsymbol{r}=H\boldsymbol{x}=\Psi\boldsymbol{a}$. Different basis have been shown to provide good results in the application of CS in US imaging, such as wavelets, waveatoms or 2D Fourier basis \cite{liebgott2012compressive}. In this paper the wavelet transform has been employed.

Once the blurred RF image, denoted by $\boldsymbol{\hat{r}}$, is recovered by solving the convex problem in (\ref{eq2}), one can restore the TRF $\boldsymbol{x}$ by minimizing: 

\begin{equation}
\underset{\boldsymbol{x}\in \mathbb{R}^{N}}{min} \alpha \left\|\boldsymbol{x}\right\|_p^p+\left\|\boldsymbol{\hat{r}}-H\boldsymbol{x}  \right \|{_{2}^{2}}
\label{Dec}
\end{equation}

The first term in (\ref{Dec}) aims at regularizing the TRF by a GGD statistical assumption, where $p$ is related to the shape parameter of the GGD. In this paper, we focus on shape parameters ranging from 1 to 2 ($1\leq p\leq 2$), allowing us to generalize the existing works in US image deconvolution mainly based on Laplacian or Gaussian statistics \cite{taxt2001two,michailovich2007blind,morin2013semi}.

\subsubsection{Proposed approach}While the sequential approach represents the most intuitive way to solve the compressive deconvolution problem, dividing a single problem into two separate subproblems will inevitably generate larger estimation errors as shown by the results in section \ref{sec3}. Therefore, we propose herein a method to solve the CS and deconvolution problem simultaneously. Similarly to  \cite{amizic2013compressive}, we formulate the reconstruction process into a constrained optimization problem explointing the relationship between  \eqref{eq2} and \eqref{Dec}.

\begin{equation}
\begin{split}
\underset{\boldsymbol{x}\in \mathbb{R}^{N},\boldsymbol{a}\in \mathbb{R}^{N}}{min} &\parallel \boldsymbol{a} \parallel_1  + \alpha \left\|\boldsymbol{x}\right\|_p^p +  \frac{1}{2\mu}\parallel \boldsymbol{y}-\Phi\Psi\boldsymbol{a} \parallel{_{2}^{2}}\\
s.t.\quad & H\boldsymbol{x}=\Psi\boldsymbol{a}
\end{split}
\label{eq3}
\end{equation}

However, since our goal is to recover enhanced US imaging by estimation the TRF $\boldsymbol{x}$, we further reformulate the problem above into a unconstrained optimization problem:

\begin{equation}
\underset{\boldsymbol{x}\in \mathbb{R}^{N}}{min} \parallel \Psi^{-1} H\boldsymbol{x} \parallel_1  + \alpha \left\|\boldsymbol{x}\right\|_p^p +  \frac{1}{2\mu}\parallel \boldsymbol{y}-\Phi H\boldsymbol{x} \parallel{_{2}^{2}}\label{eq3}
\end{equation}

The objective function in \eqref{eq3} contains, in addition to the data fidelity term, two regularization terms.  The first one aims at imposing the sparsity of the RF data $H\boldsymbol{x}$ (i.e. minimizing the  $\ell_1$-norm of the target image $\boldsymbol{x}$ convolved with a bandlimited function) in a transformed domain $\Psi$. We should note that such an assumption has been extensively used in the application of CS in US imaging, see e.g. \cite{quinsac2012frequency, liebgott2012compressive, liebgott2013pre, schiffner2014pulse, LORI-15, 7174535}. Transformations such as 2D Fourier, wavelet or wave atoms
have been shown to provide good results in US imaging. The second term aims at regularizing the TRF $\boldsymbol{x}$ and is related to the GGD statistical assumption of US images, see e.g. \cite{alessandrini2011restoration, zhao2014restoration,7174535}.

We notice that our regularized reconstruction problem based on the objective function in \eqref{eq3} is different from a typical CS reconstruction. Specifically, the objective function of a standard CS technique applied to our model would only contain the classical data fidelity term and an $\ell_1$-norm penalty similar to the first term in \eqref{eq3} but without the operator $H$. However, such a CS reconstruction is not adapted to compressive deconvolution, mainly because the requirements of CS theory such as the restricted isometry property might not be guaranteed \cite{amizic2013compressive}.

To solve the optimization problem in eq. (\ref{eq3}), we propose hereafter an algorithm based on the alternating direction method of multipliers (ADMM).

\subsection{Basics of Alternating Direction Method of Multipliers}
Before going into the details of our algorithm, we report in this paragraph the basics of ADMM. ADMM has been extensively studied in the areas of convex programming and variational inequalities, e.g., \cite{boyd2011distributed}. The general optimization problem considered in ADMM framework is as follows:

\begin{equation}
\begin{split}
\underset{u,v}{min} \qquad f(u)+g(v) \qquad\qquad   \\
s.t.\qquad  Bu + Cv = b,u\in \mathcal{U}, v\in \mathcal{V}
\end{split}
\label{eq4}
\end{equation}

where $\mathcal{U} \subseteq \mathbb{R}^s$ and $\mathcal{V} \subseteq \mathbb{R}^t$ are given convex sets, $f:\mathcal{U} \to \mathbb{R}$ and $g:\mathcal{V} \to \mathbb{R}$ are closed convex functions, $B \in \mathbb{R}^{r\times s}$ and $C \in \mathbb{R}^{r\times t}$ are given matrices and $\boldsymbol{b}\in \mathbb{R}^r$ is a given vector.

By attaching the Lagrangian multiplier $\lambda\in\mathbb{R}^{r}$ to the linear constraint, the Augmented Lagrangian (AL) function of (\ref{eq4}) is

\begin{equation}
\begin{split}
\mathcal{L}(u,v,\lambda)=f(u)+g(v) - \lambda^t(Bu + Cv - b)\\
+ \frac{\beta}{2}\parallel Bu + Cv - b\parallel_{2}^{2}
\end{split}
\label{eq5}
\end{equation}

where $\beta>0$ is the penalty parameter for the linear constraints to be satisfied. The standard ADMM framework follows the three steps iterative process:
 	
\begin{equation} 
\left\{\begin{matrix}
u^{k+1} \in \underset{u\in\mathcal{U}}{argmin} \mathcal{L}(u,v^k,\lambda^k)\\ 
v^{k+1} \in \underset{v\in\mathcal{V}}{argmin} \mathcal{L}(u^{k+1},v,\lambda^k)\\
\lambda^{k+1} =  \lambda^{k} - \beta(Bu^{k+1} + Cv^{k+1} - b)
\end{matrix}\right.
\label{eq6}
\end{equation}

The main advantage of ADMM, in addition to the relative ease of implementation, is its ability to split awkward intersections and objectives to easy subproblems, resulting into iterations comparable to those of other first-order methods. 

\subsection{Proposed ADMM parameterization for Ultrasound Compressive Deconvolution}

In this subsection, we propose an ADMM method for solving the ultrasound compressive deconvolution problem in (\ref{eq3}).

Using a trivial variable change, the minimization problem in (\ref{eq3}) can be rewritten as:

\begin{equation}
\underset{\boldsymbol{x}\in \mathbb{R}^{N}}{min} \parallel \boldsymbol{w} \parallel_1  + \alpha \left\|\boldsymbol{x}\right\|_p^p +  \frac{1}{2\mu}\parallel \boldsymbol{y}- A\boldsymbol{a} \parallel{_{2}^{2}}\label{eq7}
\end{equation}

where $\boldsymbol{a}=\Psi^{-1}H\boldsymbol{x}$, $\boldsymbol{w}=\boldsymbol{a}$ and $A = \Phi\Psi$. Let us denote $\boldsymbol{z}= \begin{bmatrix}\boldsymbol{w}\\\boldsymbol{x}\end{bmatrix}$. The reformulated problem in (\ref{eq7}) can fit the general ADMM framework in (\ref{eq4}) by choosing: $f(\boldsymbol{a})= \frac{1}{2\mu}\parallel\boldsymbol{y}-A\boldsymbol{a}\parallel_2^2$, $g(\boldsymbol{z})= \parallel \boldsymbol{w} \parallel_1 + \alpha \left\|\boldsymbol{x}\right\|_p^p$,  $B= \begin{bmatrix}I_{N}\\\Psi\end{bmatrix}$, $C=\begin{bmatrix}
-I_{N}& \boldsymbol{0}\\\boldsymbol{0}&-H
\end{bmatrix}$ and  $\boldsymbol{b}=\boldsymbol{0}$. $I_N\in\mathbb{R}^{N\times N}$ is the identity matrix.

The augmented Lagrangian function of (\ref{eq7}) is given by

\begin{equation}
\begin{split}
\mathcal{L}(\boldsymbol{a},\boldsymbol{z},\boldsymbol{\lambda})=f(\boldsymbol{a})+g(\boldsymbol{z}) - \boldsymbol{\lambda}^t(B\boldsymbol{a} + C\boldsymbol{z})\\
+ \frac{\beta}{2}\parallel B\boldsymbol{a} + C\boldsymbol{z}\parallel_{2}^{2}
\end{split}
\label{eq8}
\end{equation}

where $\boldsymbol{\lambda}\in\mathbb{R}^{2N}$ stands for $\boldsymbol{\lambda}=\begin{bmatrix}
\boldsymbol{\lambda_1}\\\boldsymbol{\lambda_2}
\end{bmatrix}$, $\boldsymbol{\lambda_i}\in\mathbb{R}^{N} (i = 1,2)$. According to the standard ADMM iterative scheme, the minimizations with respect to $\boldsymbol{a}$ and $\boldsymbol{z}$ will be performed alternatively, followed by the update of $\boldsymbol{\lambda}$. 

\subsection{Implementation Details}
In this subsection, we detail each of the three steps of our ADMM-based compressive deconvolution method. While the following mathematical developments are given for the case when the regularization term for TRF $\boldsymbol{x}$ is equal to $\parallel\boldsymbol{x}\parallel_p^p$ (adapted to US images), our approach using a generalized total variation regularization is also detailed in Appendix A and may be useful for other (medical) applications.

\textbf{Step 1} consists in solving the $\boldsymbol{z}$-problem, since $\boldsymbol{z}= \begin{bmatrix}\boldsymbol{w}\\\boldsymbol{x}\end{bmatrix}$, this problem can be further divided into two subproblems.

\textit{Step 1.1} aims at solving:

\begin{equation}
\begin{split}
\boldsymbol{w}^k=& \underset{\boldsymbol{w}\in \mathbb{R}^{N}}{argmin}\quad \parallel \boldsymbol{w}\parallel_1-(\boldsymbol{\lambda_1}^{k-1})^t(\mathbf{a}^{k-1}-\boldsymbol{w})\\
& +\frac{\beta}{2}\parallel \boldsymbol{a}^{k-1}-\boldsymbol{w}\parallel_2^2\\
\Leftrightarrow \boldsymbol{w}^k=& \underset{\boldsymbol{w}\in \mathbb{R}^{N}}{argmin} \parallel \boldsymbol{w}\parallel_1+\frac{\beta}{2} \parallel \boldsymbol{a}^{k-1} - \boldsymbol{w} - \frac{\boldsymbol{\lambda_1}^{k-1}}{\beta} \parallel{_{2}^{2}}\\
\Leftrightarrow \boldsymbol{w}^k=& prox_{\parallel\cdot\parallel_1/\beta}\left( \boldsymbol{a}^{k-1} - \frac{\boldsymbol{\lambda_1}^{k-1}}{\beta}\right)
\end{split}
\label{step1a}
\end{equation}

where $prox$ stands for the proximal operator as proposed in  \cite{pesquet2012parallel, pustelnik2011parallel, pustelnik2012relaxing}. The proximal operators of various kinds of functions including $\left\|\boldsymbol{x}\right\|_p^p$ have been given explicitly in the literature (see e.g. \cite{combettes2011proximal}). Basics about the proximal operator of $\left\|\boldsymbol{x}\right\|_p^p$ are reminded in Appendix B.

\textit{Step 1.2} consists in solving:
\begin{equation}
\begin{split}
\boldsymbol{x}^k=\underset{\boldsymbol{x}\in \mathbb{R}^{N}}{argmin}\quad \alpha \parallel\boldsymbol{x}\parallel_p^p-\boldsymbol{\lambda}_2^{k-1}(\Psi\boldsymbol{a}^{k-1}-H\boldsymbol{x})\\
+\frac{\beta}{2}\parallel\Psi\boldsymbol{a}^{k-1}-H\boldsymbol{x}\parallel_2^2
\end{split}
\label{eq19bis}
\end{equation}

For $p$ equal to $2$, the minimization in (\ref{eq19bis}) can be easily solved in the Fourier domain, as follows:

\begin{equation}
\boldsymbol{x}^{k} = \left [ \beta H^tH+2\alpha I_N\right ]^{-1}\times \left [\beta H^t\Psi \boldsymbol{a}^{k-1}-H^t\boldsymbol{\lambda}_2^{k-1} \right ]
\label{step1b2}
\end{equation}

For $1\leqslant p < 2$, we propose to use the proximal operator to solve (\ref{eq19bis}). In this case, eq. (\ref{eq19bis}) will be equivalent to 

\begin{equation}
\boldsymbol{x}^k=\underset{\boldsymbol{x}\in \mathbb{R}^{N}}{argmin}\quad \alpha \parallel \boldsymbol{x}\parallel_p^p+\frac{\beta}{2}\parallel\Psi\boldsymbol{a}^{k-1}-H\boldsymbol{x}-\frac{\boldsymbol{\lambda}_2^{k-1}}{\beta}\parallel_2^2
\label{eq23}
\end{equation}

Denoting $h(\boldsymbol{x})=\frac{1}{2} \parallel \Psi \boldsymbol{a}^{k-1} - H\boldsymbol{x} - \frac{\boldsymbol{\lambda_2}^{k-1}}{\beta} \parallel{_{2}^{2}}$, we can further approximate $h(\boldsymbol{x})$ by

\begin{equation}
h'(\boldsymbol{x}^{k-1})(\boldsymbol{x}-\boldsymbol{x}^{k-1})+\frac{1}{2\gamma}\parallel \boldsymbol{x}-\boldsymbol{x}^{k-1}\parallel_2^2
\label{eq10}
\end{equation}

where $\gamma >0$ is a parameter related to the Lipschitz constant \cite{chouzenoux2014variable} and $h'(\boldsymbol{x}^{k-1})$ is the gradient of $h(\boldsymbol{x})$ when $\boldsymbol{x}=\boldsymbol{x}^{k-1}$.

By plugging \eqref{eq10} into (\ref{eq23}), we obtain:

\begin{equation}
\begin{split}
\boldsymbol{x}^k \approx &\underset{\boldsymbol{x}\in \mathbb{R}^{N}}{argmin}\quad \alpha \parallel \boldsymbol{x}\parallel_p^p + \beta h'(\boldsymbol{x}^{k-1})(\boldsymbol{x}-\boldsymbol{x}^{k-1})\\
&+\frac{\beta}{2\gamma}\parallel \boldsymbol{x}-\boldsymbol{x}^{k-1}\parallel_2^2
\\
\Leftrightarrow\boldsymbol{x}^k\approx&\underset{\boldsymbol{x}\in \mathbb{R}^{N}}{argmin}\quad \alpha \parallel \boldsymbol{x}\parallel_p^p+\frac{\beta}{2\gamma}\parallel \boldsymbol{x}-\boldsymbol{x}^{k-1}+\gamma h'(\boldsymbol{x}^{k-1})\parallel_2^2
\end{split}
\label{eq24}
\end{equation}

According to the definition of the proximal operator, we can finally get

\begin{equation}
\boldsymbol{x}^k \approx prox_{\alpha\gamma\parallel\cdot\parallel_p^p/\beta}\{\boldsymbol{x}^{k-1}-\gamma h'(\boldsymbol{x}^{k-1})\}
\label{step1b}
\end{equation}

We should note that \eqref{step1b} provides an approximate solution, thus resulting into an inexact ADMM scheme. However, the convergence of such inexact ADMM has been already established in \cite{he2002new, boyd2011distributed, yang2011alternating}.

\textbf{Step 2} aims at solving:

\begin{equation}
\begin{split}
\boldsymbol{a}^k&=\underset{\boldsymbol{a}\in \mathbb{R}^{N}}{argmin}\quad \frac{1}{2\mu}\parallel \boldsymbol{y} - A\boldsymbol{a}\parallel_2^2-(\boldsymbol{\lambda}^{k-1})^t(B\boldsymbol{a}+C\boldsymbol{z}^k)\\
&+\frac{\beta}{2}\parallel B\boldsymbol{a}+C\boldsymbol{z}^k\parallel_2^2\\
\Leftrightarrow\boldsymbol{a}^k
&=(\frac{1}{\mu}A^tA + \beta I_N + \beta \Psi^t\Psi)^{-1}(\frac{1}{\mu}A^t\boldsymbol{y}+\boldsymbol{\lambda}_1^{k-1}+\Psi^t\boldsymbol{\lambda}_2^{k-1}\\
&+\beta\boldsymbol{w}^k+\beta\Psi^tH\boldsymbol{x}^k)
\end{split}
\label{step2}
\end{equation}

The formula above is equivalent to solving an $N\times N$ linear system or inverting an $N\times N$ matrix. However, since the sparse basis $\Psi$ considered is orthogonal (e.g. the wavelet transform), it can be reduced to solving a smaller $M\times M$ linear system or inverting an $M\times M$ matrix by exploiting the Sherman-Morrison-Woodbury inversion matrix lemma \cite{deng2013group}:

\begin{equation}
(\beta_1I_N+\beta_2A^tA)^{-1} = \frac{1}{\beta_1}I_N-\frac{\beta_2}{\beta_1}A^t(\beta_1I_M+\beta_2AA^t)^{-1}A
\label{eq12}
\end{equation}

In this paper, without loss of generality, we considered the compressive sampling matrix $\Phi$ as a Structurally Random Matrix (SRM) \cite{do2012fast}. Therefore, $A$ was formed by randomly taking a subset of rows from orthonormal transform matrices, that is, $AA^t=I_M$. As a consequence, there is no need to solve a linear system and the main computational cost consists into two matrix-vector multiplications per iteration.

\textbf{Step 3} consists in solving:

\begin{equation}
\boldsymbol{\lambda}^k = \boldsymbol{\lambda}^{k-1} - \beta(B\boldsymbol{a}^k+C\boldsymbol{z}^k)
\label{step3}
\end{equation}

The proposed optimization routine is summarized in Algorithm \ref{Algorithm:ADMM}. 

\alglanguage{pseudocode}
\begin{algorithm}
\caption{ADMM algorithm for Solving (6)}
\label{Algorithm:ADMM}

\begin{algorithmic}[1] 
	\Require $\boldsymbol{a}^0$, $\boldsymbol{\lambda}^0$, $\alpha$, $\mu$, $\beta$
 	\While{not converged}
 	\State $\boldsymbol{w}^k \gets \boldsymbol{a}^{k-1}, \boldsymbol{\lambda}^{k-1}$\Comment{update $\boldsymbol{w}^k$ using \eqref{step1a}}
 	\State $\boldsymbol{x}^k \gets \boldsymbol{a}^{k-1}, \boldsymbol{\lambda}^{k-1}$\Comment{update $\boldsymbol{x}^k$ using \eqref{step1b2} or \eqref{step1b} }
 	\State $\boldsymbol{a}^k \gets \boldsymbol{w}^k, \boldsymbol{x}^k, \boldsymbol{\lambda}^{k-1}$
\Comment{update $\boldsymbol{a}^k$ using \eqref{step2} }
 	\State $\boldsymbol{\lambda}^k \gets \boldsymbol{w}^k, \boldsymbol{x}^k, \boldsymbol{a}^k, \boldsymbol{\lambda}^{k-1}$
 	\Comment{update $\boldsymbol{\lambda}^k$ using \eqref{step3} }
    \EndWhile  
    \Ensure $\boldsymbol{x}$ 
\end{algorithmic}
\end{algorithm}

\section{Results}
\label{sec3}
The performance of the proposed compressive deconvolution method are evaluated on several simulated and experimental data sets. First, we test our algorithm on a modified Shepp-Logan phantom containing speckle noise to confirm that the $l_p$-norm regularization term is more adapted to US images than the generalized TV used in \cite{amizic2013compressive}. The approach in \cite{amizic2013compressive} is referred as CD\_Amizic hereafter. Second, we give the results of our algorithm for different $l_p$-norm optimizations on simulated US images, showing the superiority of our approach over the intuitive sequential method explained in section \ref{sec2}. Finally, compressive deconvolution results on two \textit{in vivo} ultrasound images are presented. Moreover, a comparison between our approach and CD\_Amizic on the standard Shepp-Logan phantom is provided in Appendix C.

\subsection{Results on modified Shepp-Logan phantom}

\begin{figure*}%[!t]
\begin{center}
   \includegraphics[scale=0.26]{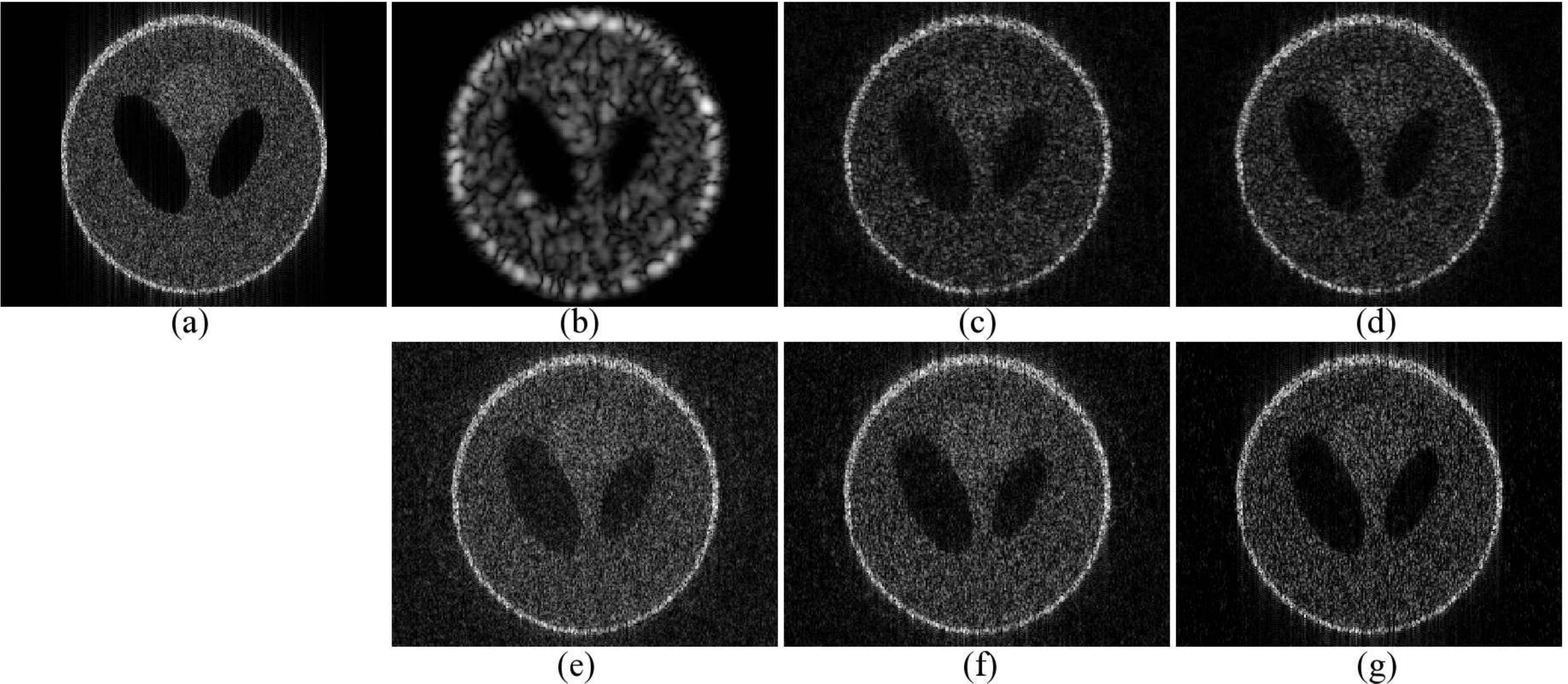}%0.5
\caption{Reconstruction results for SNR = 40dB and a CS ratio of 0.6. (a) Modified Shepp-Logan phantom containing random scatterers (TRF), (b) Degraded image by convolution with a simulated US PSF, (c) Reconstruction result with CD\_Amizic, (d) Reconstruction result with the proposed method using a generalized TV prior (ADMM\_GTV), (e, f, g) Reconstruction results with the proposed method using an $l_p$-norm prior, for $p$ equal to 1.5, 1.3 and 1 (ADMM\_L1.5, ADMM\_L1.3 and ADMM\_L1).}
\label{fig:6} 
\end{center}
\end{figure*}

We modified the Shepp-Logan phantom in order to simulate the speckle noise that degrades in practice the US images. For this, we followed the procedure classically used in US imaging \cite{ng2007wavelet}. First, scatterers at uniformly random locations have been generated, with amplitudes distributed according to a zero-mean generalized Gaussian distribution (GGD) with the shape parameter set to 1.3 and the scale parameter equal to 1. The scatterer amplitudes were further multiplied   by the values of the original Shepp-Logan phantom pixels located at the closest positions to the scatterers. The resulting image, mimicking the tissue reflectivty function (TRF) in US imaging, is shown in Fig.\ref{fig:6}(a). The blurred image in Fig.\ref{fig:6}(b) was obtained by 2D convolution between the TRF and a spatially invariant PSF generated with Field II \cite{jensen1991model}, a state-of-the-art simulator in US imaging. It corresponds to a 3.5 MHz linear probe, sampled in the axial direction at 20 MHz. The compressive measurements were obtained by projecting the blurred image onto SRM and by adding a Gaussian noise corresponding to a SNR of $40$ dB.

The results were quantitatively evaluated using the standard peak signal-to-noise ratio (PSNR) and the Structural Similarity (SSIM) \cite{wang2004image}. PSNR is defined as

\begin{equation}
PSNR = 10log_{10}\frac{NL^2}{\parallel \boldsymbol{x}-\hat{\boldsymbol{x}}\parallel^2}
\label{eq17}
\end{equation}

where $\boldsymbol{x}$ and $\hat{\boldsymbol{x}}$ are the original and reconstructed images, and the constant L represents the maximum intensity value in $\boldsymbol{x}$. 
SSIM is extensively used in US imaging and defined as

\begin{equation}
SSIM = \frac{(2\mu_x\mu_{\hat{x}}+c_1)(2\sigma_{x\hat{x}}+c_2)}{(\mu_x^2+\mu_{\hat{x}}^2+c_1)(\sigma_x^2+\sigma_{\hat{x}}^2+c_2)}
\label{eq28}
\end{equation}

where $x$ and $\hat{x}$ are the original and reconstructed images, $\mu_x$, $\mu_{\hat{x}}$, $\sigma_x$ and $\sigma_{\hat{x}}$ are the mean and variance values of $x$ and $\hat{x}$, $\sigma_{x\hat{x}}$  is the covariance between $x$ and $\hat{x}$; $c_1=(k_1L)^2$ and $c_2=(k_2L)^2$ are two variables aiming at stabilizing the division with weak denominator, $L$ is the dynamic range of the pixel-values and $k_1$, $k_2$ are constants. Herein, $L=1$, $k_1=0.01$ and $k_2=0.03$.

Reconstruction results for a CS ratio of $0.6$ are shown in Fig.\ref{fig:6}. They were obtained with: the recent compressive deconvolution technique reported in \cite{amizic2013compressive} (referred as
CD\_Amizic), the proposed method using the generalized TV prior (denoted by ADMM\_GTV) and the proposed method using the $l_p$-norm for $p$ equal to 1.5, 1.3 and 1 (denoted respectively by ADMM\_L1.5, ADMM\_L1.3 and ADMM\_L1). All the hyperparameters were set to their best possible values by cross-validation. For CD\_Amizic, $\{\beta,\alpha,\eta,\tau\}=\{10^{7},1, 10^4, 10^2\}$. For ADMM\_GTV $\{\mu,\alpha,\beta\}=\{10^{-5},2\times 10^{-1}, 10^2\}$ and for the proposed method with $l_p$-norms, $\{\mu,\alpha,\beta,\gamma\}=\{10^{-5},2\times 10^{-1}, 10^1, 3\times 10^{-2}\}$ . 
The quantitative results for different CS ratios are regrouped in Table.\ref{tab:4}. They confirm that the $l_p$-norm is better adapted to recover the TRF in US imaging than the generalized TV. The difference between the two priors is further accentuated when the CS ratio decreases.

Keeping in mind that the generalized TV prior is not well suited to model the TRF in US imaging, we did not use CD\_Amizic in the following sections dealing with simulated and experimental US images. Moreover, the proposed method was only evaluated in its $l_p$-norm minimization form.

\begin{table*}[htbp]
  \centering
  \caption{Quatitative results for the modified Shepp-Logan phantom with US speckle (SNR = 40dB)}
    \begin{tabular}{c|c|c|c|c|c|c}
    \hline
    CS ratios &      & CD\_Amizic & ADMM\_GTV & ADMM\_L1.5 & ADMM\_L1.3 & ADMM\_L1 \\
    \hline
    \multirow{2}[0]{*}{80\%}  & PSNR  & 30.82 & 31.11 & 32.23 & \textbf{32.32} & 32.05 \\
          & SSIM  & 83.24 & 85.03 & 86.44 & \textbf{88.77} & 87.70\\
    \hline
    \multirow{2}[0]{*}{60\%}  & PSNR  & 29.68 & 29.83 & 31.27 & \textbf{31.50} & 31.32 \\
          & SSIM  & 74.58 & 77.83 & 82.26 & \textbf{86.03} & 85.64 \\
    \hline
    \multirow{2}[0]{*}{40\%}  & PSNR  & 26.76 & 28.11 & 29.58 & 30.04 & \textbf{30.12} \\
          & SSIM  & 43.43 & 61.46 & 73.88 & 79.95 & \textbf{81.75} \\
    \hline
    \multirow{2}[0]{*}{20\%}  & PSNR  & 20.22 & 21.53 & 26.81 & 27.29 & \textbf{28.20} \\
          & SSIM  & 8.35  & 12.77 & 51.70  & 62.93 & \textbf{72.34} \\

    \hline
    \end{tabular}%
  \label{tab:4}%
\end{table*}%

\subsection{Results on simulated ultrasound images} 

In this section, we compared the compressive deconvolution results with our method to those obtained with a sequential approach. The latter recovers in a first step the blurred US image from the CS measurements and reconstructs in a second step the TRF by deconvolution. 

\begin{figure}%[!t]
\begin{center}
   \includegraphics[scale=0.25]{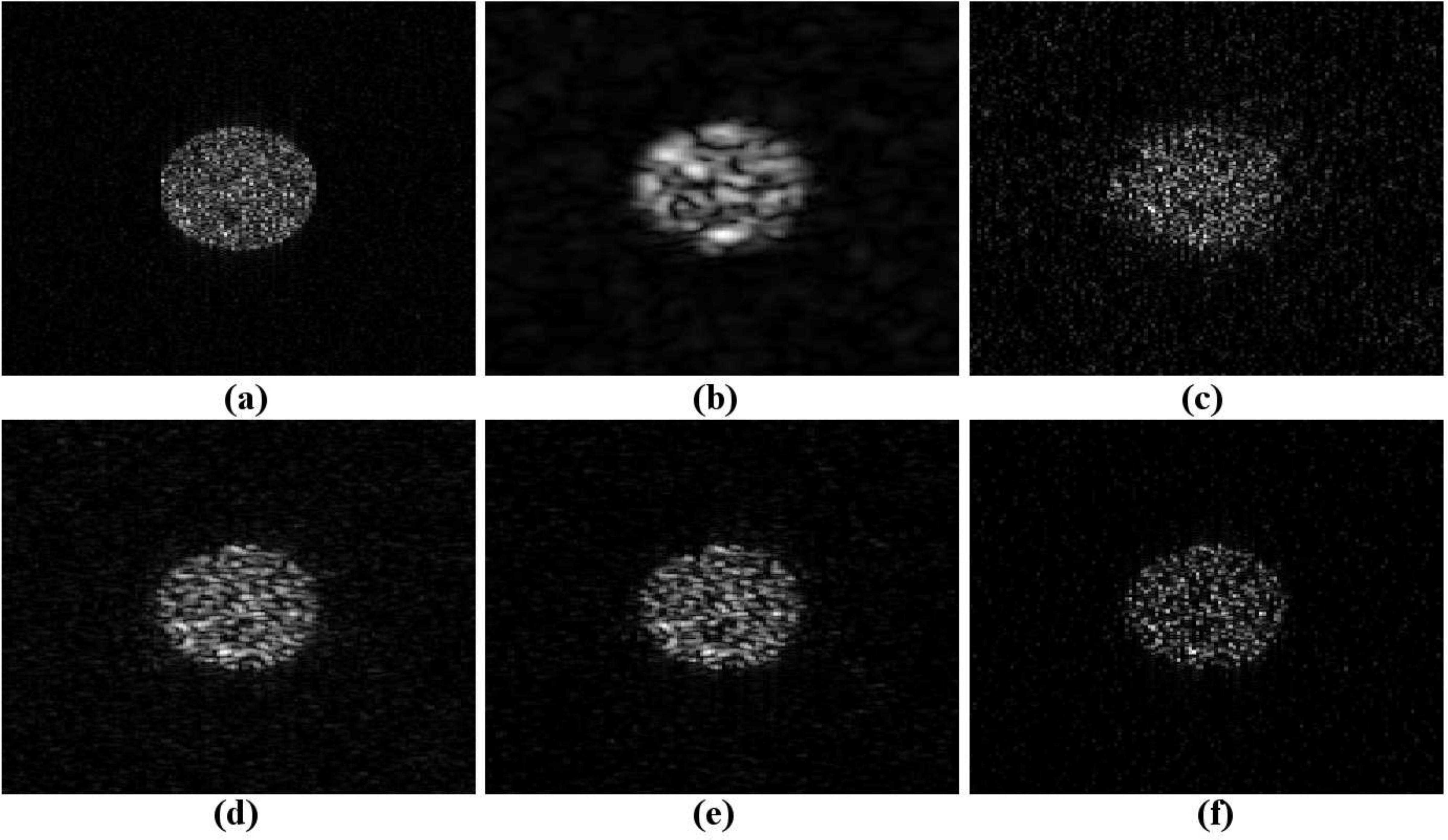}%0.5
\caption{Simulated US image and its compressive deconvolution results for a CS ratio of 0.4 and a SNR of 40 dB. (a) Original tissue reflectivity function, (b) Simulated US image, (c) Results using the sequential method, (d, e, f) Results with the proposed method for $p$ equal to 1.5, 1.3 and 1 respectively.}
\label{fig:1circle} 
\end{center}
\end{figure}

\begin{figure}%[!t]
\begin{center}
   \includegraphics[scale=0.26]{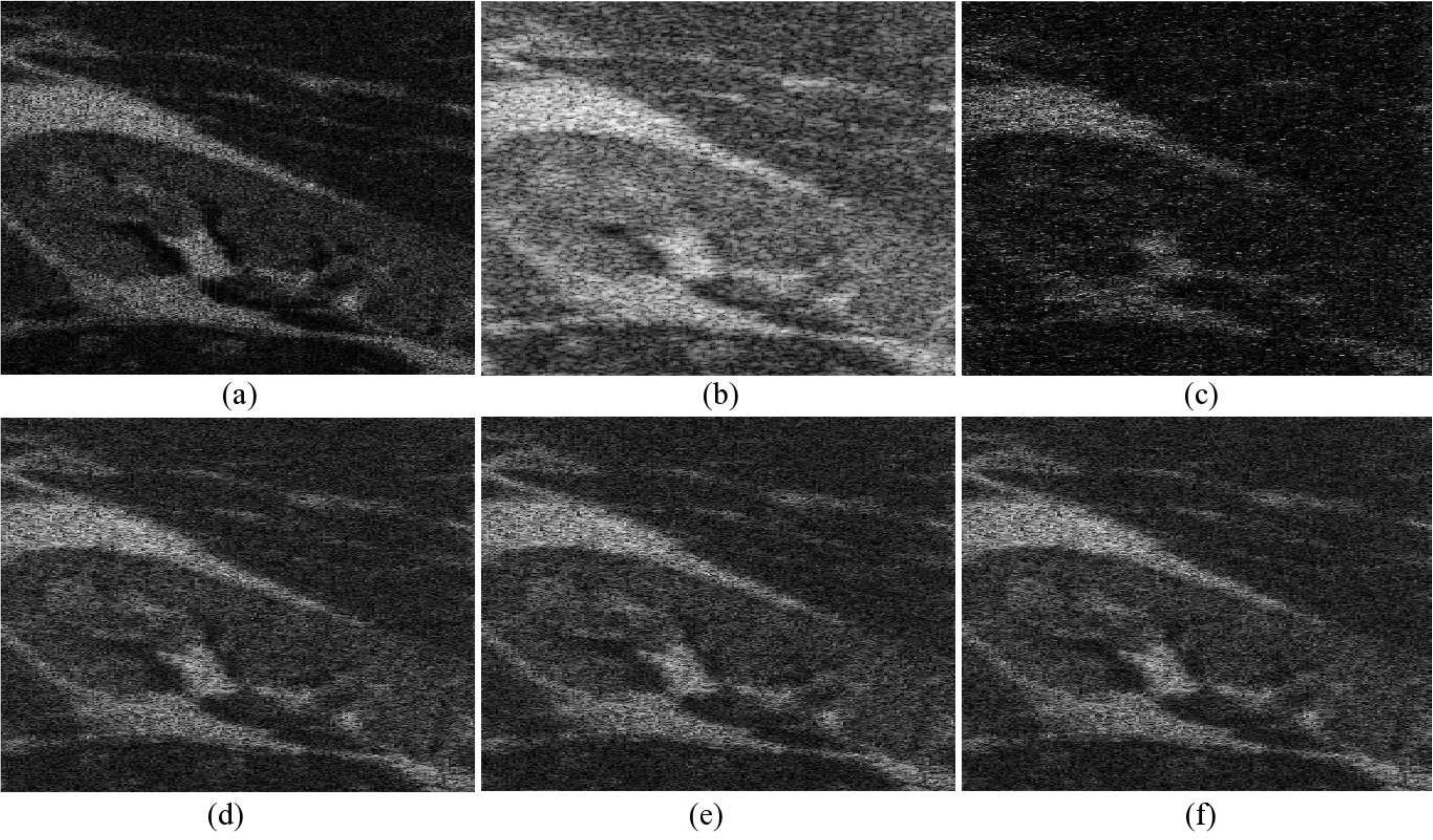}%0.5
\caption{Simulated kidney image and its compressive deconvolution results for a CS ratio of 0.2 and a SNR of 40dB. (a) Original tissue reflectivity function, (b) Simulated US image, (c) Results using the sequential method, (d, e, f) Results with the proposed method for $p$ equal to 1.5, 1.3 and 1 respectively.}
\label{fig:simukidney} 
\end{center}
\end{figure}

Two ultrasound data sets were generated, as shown in Figures \ref{fig:1circle} and \ref{fig:simukidney}. They were obtained by 2D convolution between spatially invariant PSFs and the TRFs. For the first simulated image, the same PSF as in the previous section was simulated and the TRF corresponds to a simple medium representing a round hypoechoic inclusion into a homogeneous medium. The scatterer amplitudes were random variables distributed according to a GGD with the shape parameter set to 1. The second data set is one of the examples proposed by the Field II simulator \cite{jensen1991model}, mimicking a kidney tissue. The PSF was also generated with Field II corresponding to a 4 MHz central frequency and an axial sampling frequency of 40 MHz. It corresponds to a focalized emission (the PSF was measured at the focal point) with a simulated linear probe containing 128 elements. The shape parameter of the GGD used to generate the scatterer amplitudes was set to $1.5$ and the number of scatterers was considered sufficiently large ($10^6$) to ensure fully developped speckle. In both experiments, the compressed measurements were obtained by projecting the RF images on SRM, aiming at reducing the amount of data available.

\begin{figure}%[!t]
\begin{center}
   \includegraphics[scale=0.36]{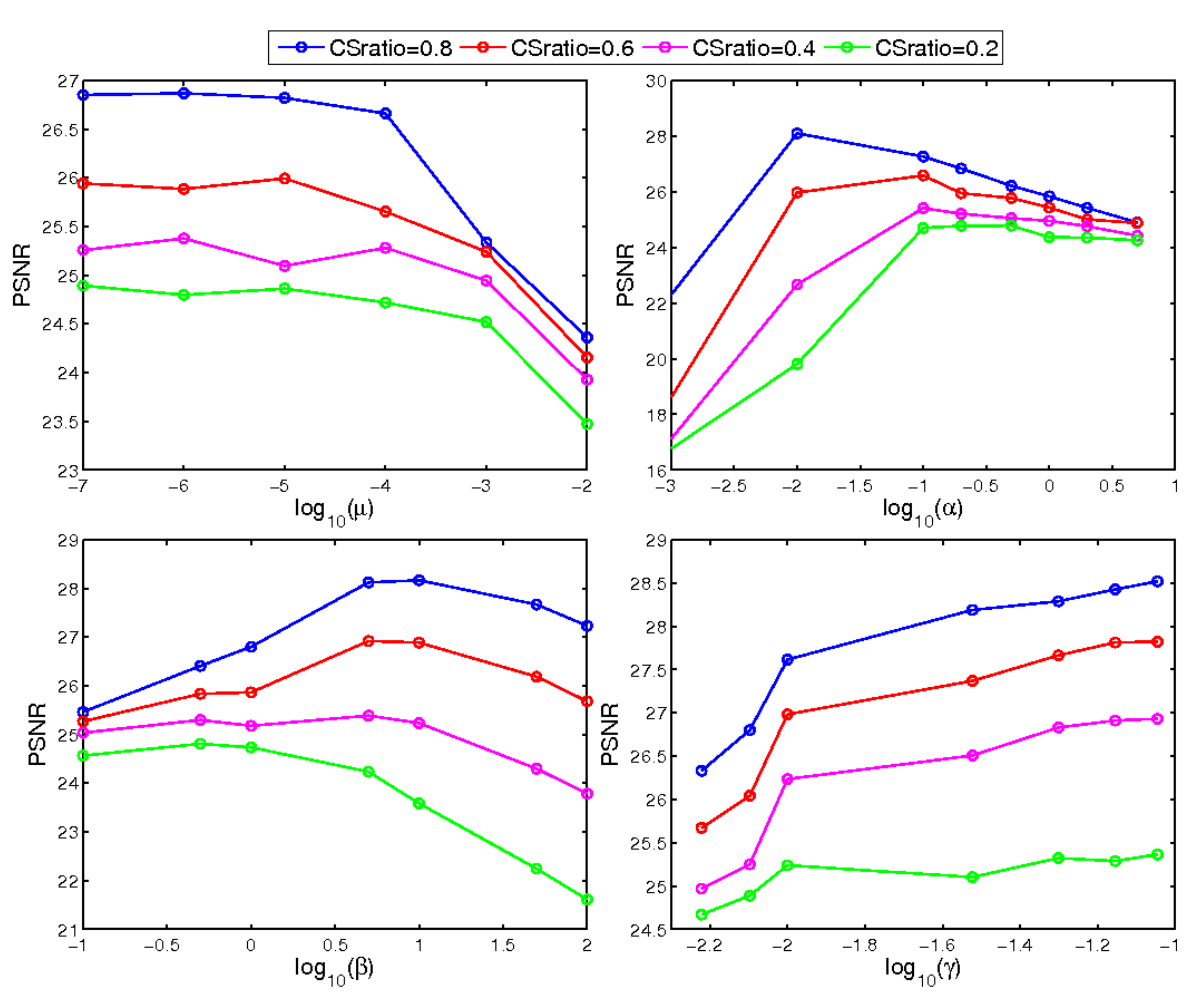}%0.5
\caption{The impact of hyperparameters on the performance of proposed algorithm on Figure. \ref{fig:1circle}.}
\label{fig:para} 
\end{center}
\end{figure} 

\begin{table}[htbp]
  \centering
  \caption{Quantitative results for simulated US images (SNR = 40dB)}
\begin{tabular}{c|c|c|c|c|c}
    \hline
%    CS Ratios &       & Sequential & Proposed ($l_{1.5}$) & Proposed ($l_{1.3}$) & Proposed ($l_1$) \\
    CS & \multirow{2}[0]{*}{} & Sequential & Proposed & Proposed & Proposed\\
    Ratios & \multirow{2}[0]{*}{} &  & ($l_{1.5}$) & ($l_{1.3}$) & ($l_{1}$)\\
    \hline
    \multicolumn{6}{c}{Figure \ref{fig:1circle}}\\
    \hline
	\multirow{2}[0]{*}{80\%} & PSNR  & 26.50 & 24.74 & 25.29 & \textbf{26.82} \\
          & SSIM  & 75.01 & 73.91 & 77.66 & \textbf{79.45} \\
    \hline
    \multirow{2}[0]{*}{60\%}  & PSNR  & 25.96 & 24.44 & 24.74 & \textbf{26.03} \\
          & SSIM  & 68.59 & 69.37 & 74.72 & \textbf{76.26} \\
    \hline
    \multirow{2}[0]{*}{40\%}  & PSNR  & 23.38 & 24.21 & 24.57 & \textbf{25.28} \\
          & SSIM  & 47.60  & 62.58 & 71.86 & \textbf{72.78} \\
    \hline
    \multirow{2}[0]{*}{20\%}  & PSNR  & 21.10  & 23.72 & 24.42 & \textbf{24.77} \\
          & SSIM  & 36.07 & 50.34 & 66.48 & \textbf{70.44} \\
    \hline
    \multicolumn{6}{c}{Figure \ref{fig:simukidney}}\\
    \hline
    	\multirow{2}[0]{*}{80\%}  & PSNR  & 26.06 & 26.71 & \textbf{26.72} & 26.69 \\
    & SSIM  & 45.99 & 56.81 & \textbf{56.84} & 56.71 \\
    \hline
    \multirow{2}[0]{*}{60\%}  & PSNR  & 25.44 & \textbf{26.38} & 26.31 & 26.29 \\
    & SSIM  & 38.86 & \textbf{54.14} & 53.90 & 53.80 \\
    \hline
    \multirow{2}[0]{*}{40\%}  & PSNR  & 25.37 & 25.89 & 25.95 & \textbf{25,97} \\
    & SSIM  & 34.61 & 50.22 & 50.51 & \textbf{50.61} \\
    \hline
    \multirow{2}[0]{*}{20\%}  & PSNR  & 24.96 & \textbf{25.22} & 25.20  & 25.12 \\
    & SSIM  & 30.89 & \textbf{41.41} & 41.32 & 40.97 \\

    \hline
        \end{tabular}%
  \label{tab:simu}%
\end{table}%

With the sequential approach, YALL1 \cite{yang2011alternating} was used to process the CS reconstruction following the minimization in eq. (\ref{eq2}). The deconvolution step was processed using the Forward-Backward Splitting method \cite{combettes2005signal, raguet2013generalized}. Both the CS reconstruction and the deconvolution procedures were performed with the same priors as the proposed compressive deconvolution approach.

The algorithm stops when the convergence criterion $\parallel \boldsymbol{x}^k - \boldsymbol{x}^{k-1} \parallel/\parallel \boldsymbol{x}^{k-1}\parallel<1e^{-3} $ is satisfied. %The hyperparameters were set to their best possible values by cross-validation. 
In order to highlight the influence of %$\alpha$ and $\beta$ 
these hyperparameters on the reconstruction results, we consider the simulated US image in Fig. \ref{fig:1circle}. The PSNR values obtained while varying the values of these hyperparameters are shown in Fig. \ref{fig:para}. From Fig. \ref{fig:para}, one can observe that the best results are obtained for small values of $\mu$, corresponding to an important weight given to the data attachment term. The best value of $\alpha$ is the one providing the best compromise between the two prior terms considered in eq. \eqref{eq3}, promoting minimal $\ell_1$-norm of $H\boldsymbol{x}$ in the wavelet domain and GGD statistics for $\boldsymbol{x}$. The choice of $\beta$ and $\gamma$ parameters, used in the augmented Lagrangian function and in the approximation of the $\ell_p$-norm proximal operator, have an important impact on the algorithm convergence. Moreover, one may observe that for a given range of values, the choice of $\gamma$ has less impact on the quality of the results than the other three hyperparameters. Despite different optimal values for each CS ratio, in the results presented through the paper, we considered their values fixed for all the CS ratios. The hyperparameters with our approach were set to $\{\mu,\alpha,\beta,\gamma\}=\{10^{-5},2\times 10^{-1}, 1, 10^{-2}\}$ for the round cyst image and $\{\mu,\alpha,\beta,\gamma\}=\{10^{-5},2\times 10^{-1}, 1\times 10^{3}, 10^{-4}\}$ for the simulated kidney image.

The quantitative results in Table \ref{tab:simu} show that the proposed method outperforms the sequential approach, for all the CS ratios and values of $p$ considered. They confirm the visual impression given by Figures \ref{fig:1circle} and \ref{fig:simukidney}. We should remark that for the first simulated data set, the $l_1$-norm gives the best result. This may be explained by the simple geometry of the simulated TRF, namely its sparse appearance. The second data set, more realistic and more representative of experimental situations, shows the interest of using different values of $p$. It confirms the generality interest of the proposed method, namely its flexibility in the choice of TRF priors. 

\subsection{In vivo study}

\begin{figure*}%[!t]
\begin{center}
   \includegraphics[scale=0.25]{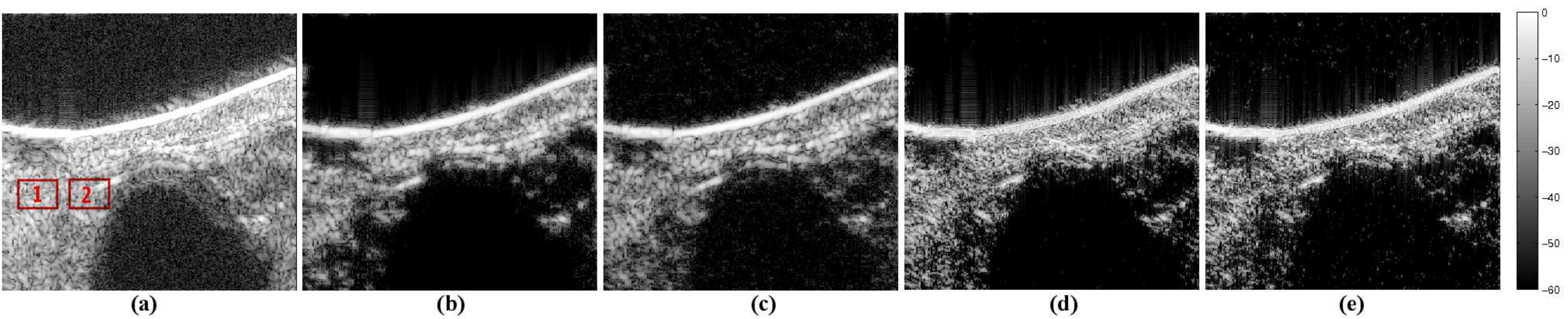}%0.5
\caption{From left to right, the original \textit{in vivo} image and its compressive deconvolution results for CS ratios of 1, 0.8, 0.6 and 0.4 respectively with $p=1$.}
\label{fig:bladder} 
\end{center}
\end{figure*}

\begin{figure*}%[!t]
\begin{center}
   \includegraphics[scale=0.25]{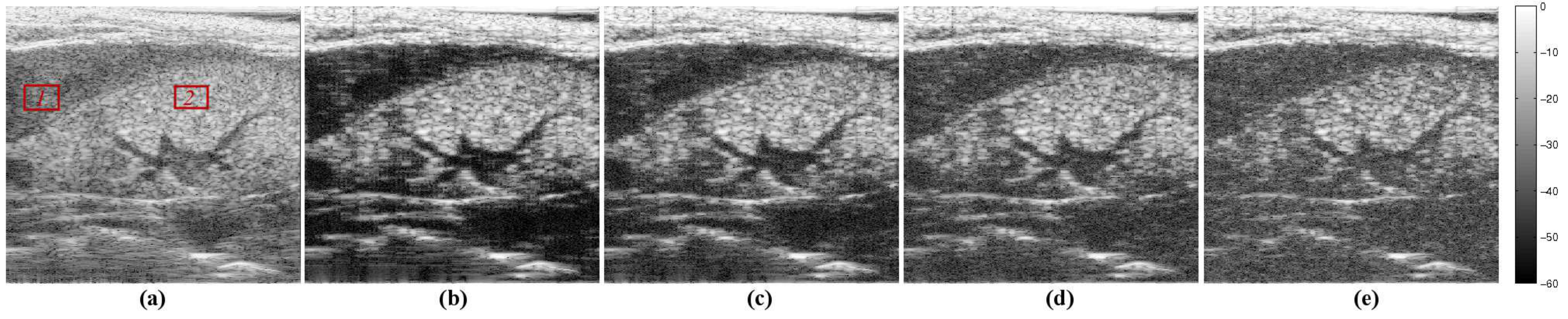}%0.5
\caption{From left to right, the original \textit{in vivo} image and its compressive deconvolution results for CS ratios of 1, 0.8, 0.6 and 0.4 respectively with $p=1.5$.}
\label{fig:kidney} 
\end{center}
\end{figure*}

In this section, we tested our method with two \textit{in vivo} data sets. The experimental data were acquired with a 20 MHz single-element US probe on a mouse bladder (first example) and kidney (second example). Unlike the simulated cases studied previously, the PSF is not known in these experiments and has to be estimated from the data. In this paper, the PSF estimation method presented in \cite{michailovich2005novel} has been adopted. The PSF estimation adopted is not iterative and the computational time for this pre-processing step is negligible compared to the reconstruction process. The compressive deconvolution results are shown in Figures \ref{fig:bladder} and \ref{fig:kidney} for different CS ratios.   

Given that the true TRF is not known in experimental conditions, the quality of the reconstruction results is evaluated using the contrast-to-noise ratio (CNR) \cite{lyshchik2005elastic}, defined as

\begin{equation}
CNR = \frac{\left|\mu_1-\mu_2\right|}{\sqrt{\sigma_1^2+\sigma_2^2}}
\end{equation}

where $\mu_1$ and $\mu_2$ are the mean of pixels located in two regions extracted from the image while $\sigma_1$ and $\sigma_2$ are the standard deviations of the same blocks. The two regions selected for the computation of the CNR are highlighted by the two red rectangles in Figures \ref{fig:bladder}(a) and \ref{fig:kidney}(a). Table. \ref{invivo} gives the CNR assessment for these two \textit{in vivo} data sets with different CS ratios and $p$ values. Given the sparse appearance of the bladder image in Fig. \ref{fig:bladder}(a), the best result was obtained for $p$ equal to $1$. However, the complexity of the tissue structures in the kidney image in Fig. \ref{fig:kidney} results into better results for $p$ larger than $1$. Nevertheless, both the visual impression and the CNR results show the ability of our method to both recover the image from compressive measurements and to improve its contrast compared to the standard US image. In particular, we may remark the improved contrast of the structures inside the kidney on our reconstructed images compared to the original one.

\begin{table}[htbp]
  \centering
  \caption{CNR assessment for \textit{in vivo} data}
    \begin{tabular}{c|c|c|cccc}
    \hline
    \multicolumn{1}{c|}{\multirow{2}[0]{*}{Figure}} & \multicolumn{1}{c|}{\multirow{2}[0]{*}{Original CNR}} & \multicolumn{1}{c|}{\multirow{2}[0]{*}{$p$ values}} & \multicolumn{4}{c}{CS ratios} \\
    \multicolumn{1}{c|}{} & \multicolumn{1}{c|}{} & \multicolumn{1}{c|}{} & 100\% & 80\%  & 60\%  & 40\% \\
    \hline
    \multirow{2}[0]{*}{Fig.6} & \multirow{2}[0]{*}{1.106} & p = 1   & \textbf{1.748} & \textbf{1.546} & \textbf{1.367} & \textbf{1.333} \\
          &       & p = 1.5 &  1.690 & 1.424   &  1.304   & 1.287 \\
    \hline
    \multirow{2}[0]{*}{Fig.7} & \multirow{2}[0]{*}{1.316} & p = 1   & \textbf{2.373} & \textbf{2.162}   & 1.895   & 1.434  \\
          &       & p = 1.5 & 2.317 & 2.082 & \textbf{1.905} & \textbf{1.451} \\
    \hline
    \end{tabular}%
	\label{invivo}%
\end{table}%

\section{Conclusion}
\label{sec4}
This paper introduced an ADMM-based compressive deconvolution framework for ultrasound imaging systems. The main benefit of our approach is its ability to reconstruct enhanced ultrasound RF images from compressed measurements, by inverting a linear model combining random projections and 2D convolution. Compared to a standard compressive sampling reconstruction that operates in the sparse domain, our method solves a regularized inverse problem combining the data attachment and two prior terms. One of the regularizers promotes minimal $\ell_1$-norm of the target image transformed by 2D convolution with a bandlimited ultrasound PSF. The second one is seeking for imposing GGD statistics on the tissue reflectivity function to be reconstructed. 
Simulation results in Appendix C on the standard Shepp-Logan phantom show the superiority of our method, both in accuracy and in computational time, over a recently published compressive deconvolution approach. Moreover, we show that the proposed joint CS and deconvolution approach is more robust than an intuitive technique consisting of first reconstructing the RF data and second deconvolving it. Finally, promising results on \textit{in vivo} data demonstrate the effectiveness of our approach in practical situations. We emphasize that the 2D convolution model may not be valid over the entire image because of the spatially variant PSF. While in this paper we focused on compressive image deconvolution based on  spatially invariant PSF, a more complicated global
model combining several local shift invariant PSFs represents
an interesting perspective of our approach. 
Our future work includes: I) the consideration of blind deconvolution techniques able to estimate (update) the spatially variant or invariant PSF during the reconstruction process, II) automatic techniques for choosing the optimal parameter $p$ used to regularize the tissue reflectivity function, III) extend our method to 3D ultrasound imaging, IV) evaluate other existing setups to obtain the random compressed measurements further adapted to accelerate the frame rate instead of only reducing the amount of acquired data, V) consider the compressed deconvolution of temporal image sequences by taking advantage of the information redundancy and by including in our model the PSF frame-to-frame variation caused by strong clutters in \textit{in vivo} scenarios, VI) evaluate our approach on more experimental data.

\appendices
\section{Proposed compressive deconvolution with generalized TV}
Although the generalized total variation (TV) used in \cite{amizic2013compressive} is not suitable for ultrasound images, it may have great interest in other (medical) application dealing with piecewise constant images. As suggested in \cite{amizic2013compressive}, the generalized TV is given by:

\begin{equation}
\sum_{d\in D}2^{1-o(d)}\sum_{i}\left |\Delta _i^d(\boldsymbol{x})  \right |^p\label{eq15}
\end{equation}

where $o(d)\in\{1,2\}$ denotes the order of the difference operator $\Delta_i^d(\boldsymbol{x})$, $0<p<1$, and $d\in D=\{h,v,hh,vv,hv\}$. $\Delta_i^h(\boldsymbol{x})$ and $\Delta_i^v(\boldsymbol{x})$ correspond, respectively, to the horizontal and vertical first order differences, at pixel $i$, that is, $\Delta_i^h(\boldsymbol{x})=u_i-u_{l(i)}$ and $\Delta_i^v(\boldsymbol{x})=u_i-u_{a(i)}$, where $l(i)$ and $a(i)$ denote the nearest neighbors of $i$, to the left and above, respectively. The operators $\Delta_i^{hh}(\boldsymbol{x})$, $\Delta_i^{vv}(\boldsymbol{x})$, $\Delta_i^{hv}(\boldsymbol{x})$ correspond, respectively, to horizontal, vertical and horizontal-vertical second order differences, at pixel $i$.

Replacing the $\ell_p$-norm by the generalized TV in our compressive deconvolution scheme results in a modified $\boldsymbol{x}$ update step, that turns in solving:

\[
\begin{split}
\boldsymbol{x}^k=&\underset{\boldsymbol{x}\in \mathbb{R}^{N}}{argmin}\quad \alpha \sum_{d\in D}2^{1-o(d)}\sum_{i}\left |\Delta _i^d(\boldsymbol{x})  \right |^p\\
&-\boldsymbol{\lambda}_2^{k-1}(\Psi\boldsymbol{a}^{k-1}-H\boldsymbol{x})
+\frac{\beta}{2}\parallel\Psi\boldsymbol{a}^{k-1}-H\boldsymbol{x}\parallel_2^2
\end{split}
\]

Similarly to the first step of the method in \cite{amizic2013compressive}, the equation above can be solved iteratively by:

\begin{equation}
\begin{split}
\boldsymbol{x}^{k,l} = &\left [ \beta H^tH+\alpha p \sum_d2^{1-o(d)}(\boldsymbol{\Delta}^d)^tB_d^{k,l}(\boldsymbol{\Delta}^d) \right ]^{-1}\\
&\times \left [\beta H^t\Psi \boldsymbol{a}^{k-1}-H^t\boldsymbol{\lambda}_2^{k-1} \right ]
\end{split}
\label{step1b1}
\end{equation}

where $l$ is the iteration number in the process of updating $\boldsymbol{x}$, $B_d^{k,l}$ is a diagonal matrix with entries $\boldsymbol{\Delta}^d$ is the convolution matrix (BCCB matrix) of the difference operator $\Delta_i^d(\cdot)$ and $B_d^{k,l}(i,i)=(v_{d,i}^{k,l})$, which is updated iteratively by:

\begin{equation}
v_{d,i}^{k,l+1} = [\Delta_i^d(\boldsymbol{x}^{k,l})]^2
\label{step1b3}
\end{equation}

When a stopping criterion is met, we can finally get an update of $\boldsymbol{x}$. 

\section{Proximal Operator}

The proximal operator of a function $f$ is defined for $x^0\in\mathbb{R}^N$ by:

\begin{equation}
prox_f(x^0) = \underset{\boldsymbol{x}\in \mathbb{R}^{N}}{argmin}\quad f(x)+\frac{1}{2}\parallel x-x^0\parallel_2^2
\label{eq20}
\end{equation}

When $f=K|x|^p$, the corresponding porximal operator has been given by \cite{combettes2011proximal}:

\begin{equation}
prox_{K\left |x\right|^p}(x^0)=sign(x^0)q
\label{eq26}
\end{equation}

where $q\geqslant 0$ and 

\begin{equation}
q+pKq^{p-1}=\left |x^0\right |
\label{eq27}
\end{equation}

It is obvious that the proximal operator of $K\left |x\right |$ is a soft thresholding, which is equal to:

\begin{equation}
prox_{K\left |x\right|}(x^0) = max\left\{\left |x^0\right |- K,0\right\}\frac{x^0}{\left|x^0\right |}
\label{eq21}
\end{equation}

When $p\neq 1$, we used Newton's method to obtain its numerical solution, \textit{i.e.} the value of $q$.

\section{Comparison with the compressive deconvolution method in \cite{amizic2013compressive}}

In this appendix we show an experiment aiming to evaluate the performance of the proposed approach compared to the one in \cite{amizic2013compressive}, denoted by \ CD\_Amizic herein. The comparison results are obtained on the standard $256\times 256$ Shepp-Logan phantom. The measurements have been generated in a similar manner as in \cite{amizic2013compressive}, \textit{i.e.} the original image was normalized, degraded by a $2D$ Gaussian PSF with a 5-pixel variance, projected onto a structured random matrix (SRM) to generate the CS measurements and corrupted by an additive Gaussian noise. We should remark that in \cite{amizic2013compressive} the compressed measurements were originally generated using a Gaussian random matrix. However, we have found that the reconstruction results with CD\_Amizic are slightly better when using a SRM compared to the PSNR results reported in [26]. Both methods were based on the generalized TV to model the image to be estimated and the 3-level Haar wavelet transform as sparsifying basis $\Psi$. With our method, the hyperparameters were set to $\{\alpha,\mu,\beta\}=\{10^{-1}, 10^{-5},10^2\}$. The same hyperparameters as reported in \cite{amizic2013compressive} were used for CD\_Amizic. Both algorithms based on the non-blind deconvolution (PSF is supposed to be known) and used the same stopping criteria.

\begin{figure*}
\begin{center}
   \includegraphics[scale=0.26]{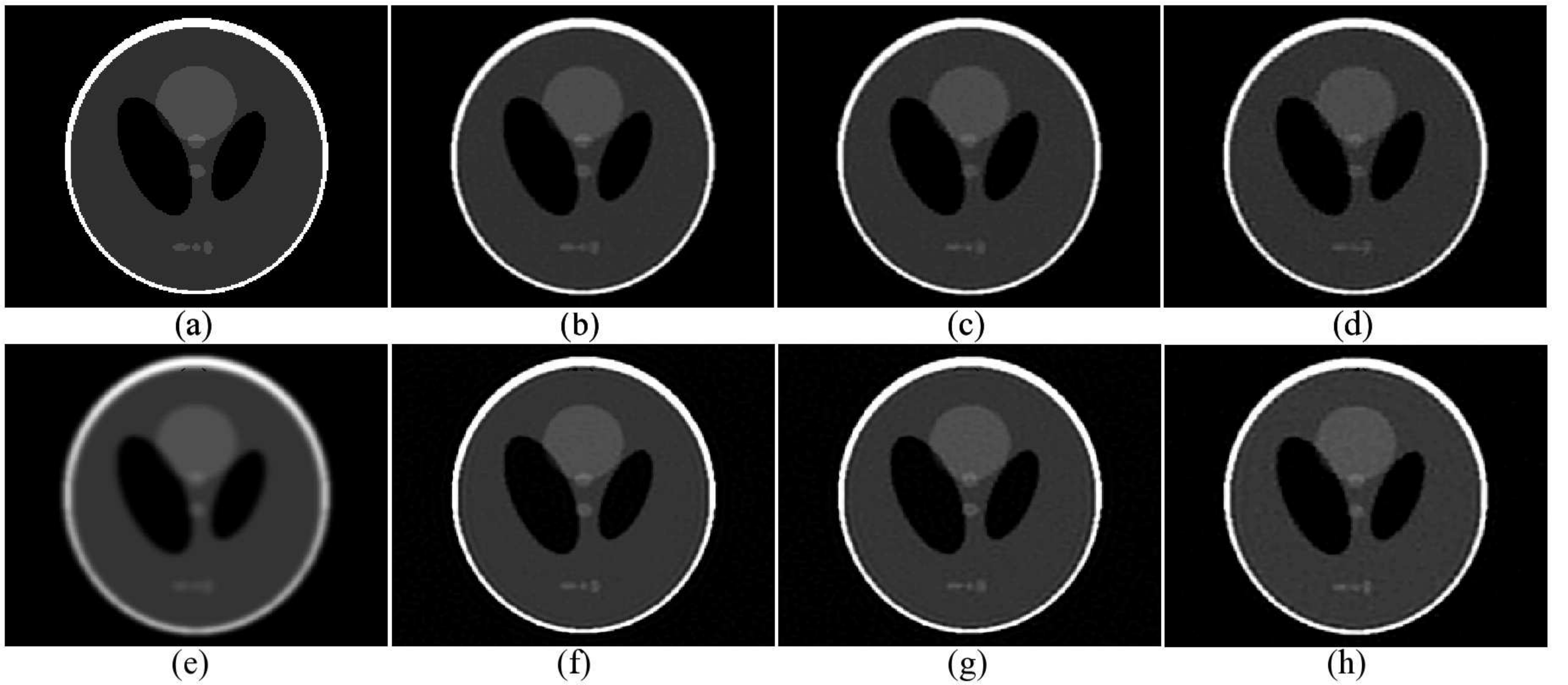}%0.5
\caption{Shepp-logan image and its compressive deconvolution results for a SNR of 40dB. (a) Original image, (e) Blurred image, (b,c,d) Compressive deconvolution results with CD\_Amizic for CS ratios of 0.8, 0.6 and 0.4, (f,g,h) Compressive deconvolution results with the proposed method for CS ratios of 0.8, 0.6 and 0.4.}
\label{fig:1} 
\end{center}
\end{figure*}

\begin{figure}
\begin{center}
   \includegraphics[scale=0.6]{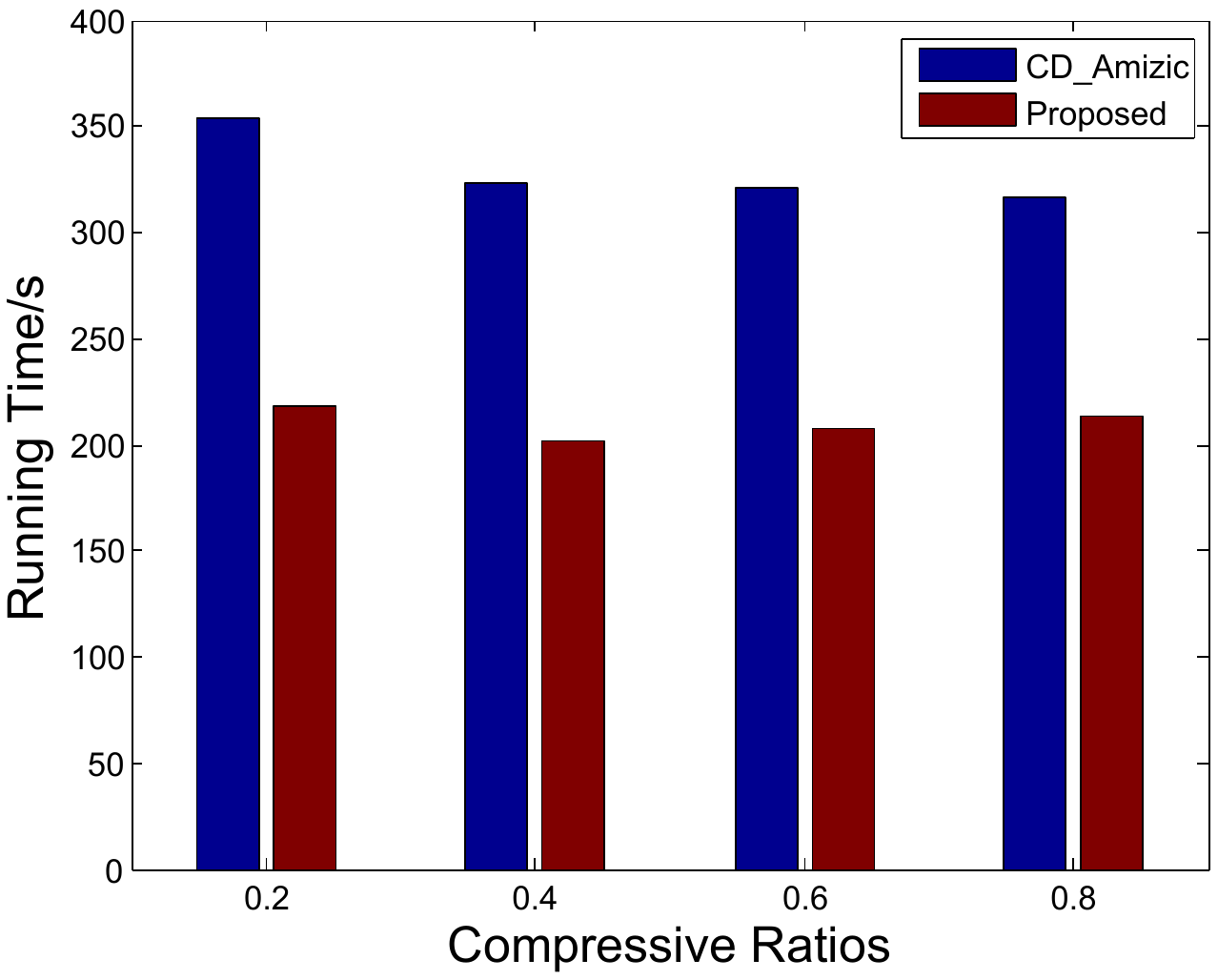}%0.5
\caption{Mean reconstruction running time for 10 experiments conducted for each CS ratio for a SNR of $40$ dB.}
\label{fig:2} 
\end{center}
\end{figure}

Fig.\ref{fig:1} shows the original Shepp-Logan image, its blurred version and a series of compressive deconvolution reconstructions using both our method and CD\_Amizic for CS ratios running from $0.4$ to $0.8$ and a SNR of $40$ dB.
Table.\ref{tab:1} regroups the PSNRs obtained with our method and with CD\_Amizic for two SNRs and for four CS ratios from $0.2$ to $0.4$. In each case, the reported PSNRs are the mean values of $10$ experiments. We may observe that our method outperforms CD\_Amizic in all the cases, allowing a PSNR improvement in the range of $0.5$ to
$2$ dB. Moreover, Fig.\ref{fig:2} shows the computational times with CD\_Amizic and the proposed method, obtained with Matlab implementations (for CD\_Amizic, the original code provided by the authors of \cite{amizic2013compressive} has been employed on a standard desktop computer (Intel Xeon CPU E5620 @ 2.40GHz, 4.00G RAM). We notice that our approach is less time consuming than CD\_Amizic for all the CS ratios considered.

\begin{table}[htbp]
  \centering
  \caption{PSNR assessment for Shepp-Logan phantom}
    \begin{tabular}{c|c|c|c|c|c}
    \hline
      SNR    & CS ratios & 20\%  & 40\%  & 60\%  & 80\% \\
    \hline
    \multirow{2}{*}{40dB}  & CD\_Amizic & 23.04 & 24.88 & 25.30 & 25.51 \\
          & Proposed method & \textbf{24.09} & \textbf{25.38} & \textbf{26.26} & \textbf{26.91} \\
    \hline
    \multirow{2}{*}{30dB}  & CD\_Amizic & 22.61  & 24.05 & 24.40 & 24.55\\
          & Proposed method &  \textbf{23.92} & \textbf{25.12} & \textbf{25.82} & \textbf{26.33} \\
    \hline
    \end{tabular}%
  \label{tab:1}%
\end{table}%

\section*{Acknowledgment}
The authors would like to thank Prof. Rafael Molina for providing the compressive deconvolution code used for comparison purpose in this paper. This work was partially supported by ANR-11-LABX-0040-CIMI within the program ANR-11-IDEX-0002-02 of the University of Toulouse and CSC (Chinese Scholarship Council).

\small
\bibliographystyle{IEEEtran}
\bibliography{CsDecADMM}

\begin{thebibliography}{10}
\providecommand{\url}[1]{#1}
\csname url@samestyle\endcsname
\providecommand{\newblock}{\relax}
\providecommand{\bibinfo}[2]{#2}
\providecommand{\BIBentrySTDinterwordspacing}{\spaceskip=0pt\relax}
\providecommand{\BIBentryALTinterwordstretchfactor}{4}
\providecommand{\BIBentryALTinterwordspacing}{\spaceskip=\fontdimen2\font plus
\BIBentryALTinterwordstretchfactor\fontdimen3\font minus
  \fontdimen4\font\relax}
\providecommand{\BIBforeignlanguage}[2]{{%
\expandafter\ifx\csname l@#1\endcsname\relax
\typeout{** WARNING: IEEEtran.bst: No hyphenation pattern has been}%
\typeout{** loaded for the language `#1'. Using the pattern for}%
\typeout{** the default language instead.}%
\else
\language=\csname l@#1\endcsname
\fi
#2}}
\providecommand{\BIBdecl}{\relax}
\BIBdecl

\bibitem{szabo2004diagnostic}
T.~L. Szabo, \emph{Diagnostic ultrasound imaging: inside out}.\hskip 1em plus
  0.5em minus 0.4em\relax Academic Press, 2004.

\bibitem{tanter2014ultrafast}
M.~Tanter and M.~Fink, ``Ultrafast imaging in biomedical ultrasound,''
  \emph{Ultrasonics, Ferroelectrics, and Frequency Control, IEEE Transactions
  on}, vol.~61, no.~1, pp. 102--119, January 2014.

\bibitem{achim2010compressive}
A.~Achim, B.~Buxton, G.~Tzagkarakis, and P.~Tsakalides, ``Compressive sensing
  for ultrasound rf echoes using a-stable distributions,'' in \emph{Engineering
  in Medicine and Biology Society (EMBC), 2010 Annual International Conference
  of the IEEE}.\hskip 1em plus 0.5em minus 0.4em\relax IEEE, 2010, pp.
  4304--4307.

\bibitem{quinsac2012frequency}
\BIBentryALTinterwordspacing
C.~Quinsac, A.~Basarab, and D.~Kouam{\'e}, ``Frequency domain compressive
  sampling for ultrasound imaging,'' \emph{Advances in Acoustics and
  Vibration}, vol. 2012, 2012. [Online]. Available:
  \url{http://dx.doi.org/10.1155/2012/231317}
\BIBentrySTDinterwordspacing

\bibitem{chernyakova2014fourier}
T.~Chernyakova and Y.~C. Eldar, ``Fourier-domain beamforming: the path to
  compressed ultrasound imaging,'' \emph{Ultrasonics, Ferroelectrics, and
  Frequency Control, IEEE Transactions on}, vol.~61, no.~8, pp. 1252--1267,
  2014.

\bibitem{liebgott2012compressive}
H.~Liebgott, A.~Basarab, D.~Kouam{\'e}, O.~Bernard, and D.~Friboulet,
  ``Compressive sensing in medical ultrasound,'' in \emph{Ultrasonics Symposium
  (IUS), 2012 IEEE International}.\hskip 1em plus 0.5em minus 0.4em\relax IEEE,
  2012, pp. 1--6.

\bibitem{liebgott2013pre}
H.~Liebgott, R.~Prost, and D.~Friboulet, ``Pre-beamformed \uppercase{RF} signal
  reconstruction in medical ultrasound using compressive sensing,''
  \emph{Ultrasonics}, vol.~53, no.~2, pp. 525--533, 2013.

\bibitem{schiffner2014pulse}
M.~F. Schiffner and G.~Schmitz, ``Pulse-echo ultrasound imaging combining
  compressed sensing and the fast multipole method,'' in \emph{Ultrasonics
  Symposium (IUS), 2014 IEEE International}.\hskip 1em plus 0.5em minus
  0.4em\relax IEEE, 2014, pp. 2205--2208.

\bibitem{RICH-11}
J.~Richy, H.~Liebgott, R.~Prost, and D.~Friboulet, ``Blood velocity estimation
  using compressed sensing,'' in \emph{IEEE International Ultrasonics
  Symposium}, Orlando (USA), 2011, pp. 1427--1430.

\bibitem{donoho2006compressed}
D.~L. Donoho, ``Compressed sensing,'' \emph{Information Theory, IEEE
  Transactions on}, vol.~52, no.~4, pp. 1289--1306, 2006.

\bibitem{candes2006robust}
E.~J. Cand{\`e}s, J.~Romberg, and T.~Tao, ``Robust uncertainty principles:
  Exact signal reconstruction from highly incomplete frequency information,''
  \emph{Information Theory, IEEE Transactions on}, vol.~52, no.~2, pp.
  489--509, 2006.

\bibitem{candes2008introduction}
E.~J. Cand{\`e}s and M.~B. Wakin, ``An introduction to compressive sampling,''
  \emph{Signal Processing Magazine, IEEE}, vol.~25, no.~2, pp. 21--30, 2008.

\bibitem{LORI-15}
O.~Lorintiu, H.~Liebgott, M.~Alessandrini, O.~Bernard, and D.~Friboulet,
  ``Compressed sensing reconstruction of \uppercase{3D} ultrasound data using
  dictionary learning and line-wise subsampling,'' \emph{IEEE Transactions on
  Medical Imaging}, vol. accepted, 2015.

\bibitem{dobigeon2012regularized}
N.~Dobigeon, A.~Basarab, D.~Kouam{\'e}, and J.-Y. Tourneret, ``Regularized
  bayesian compressed sensing in ultrasound imaging,'' in \emph{Signal
  Processing Conference (EUSIPCO), 2012 Proceedings of the 20th
  European}.\hskip 1em plus 0.5em minus 0.4em\relax IEEE, 2012, pp. 2600--2604.

\bibitem{basarab2014medical}
A.~Basarab, A.~Achim, and D.~Kouam{\'e}, ``Medical ultrasound image
  reconstruction using compressive sampling and lp-norm minimization,'' in
  \emph{SPIE Medical Imaging}.\hskip 1em plus 0.5em minus 0.4em\relax
  International Society for Optics and Photonics, 2014, pp. 90\,401H--90\,401H.

\bibitem{taxt2001two}
T.~Taxt and J.~Strand, ``Two-dimensional noise-robust blind deconvolution of
  ultrasound images,'' \emph{Ultrasonics, Ferroelectrics and Frequency Control,
  IEEE Transactions on}, vol.~48, no.~4, pp. 861--866, 2001.

\bibitem{michailovich2007blind}
O.~Michailovich and A.~Tannenbaum, ``Blind deconvolution of medical ultrasound
  images: A parametric inverse filtering approach,'' \emph{IEEE Transactions on
  Image Processing}, vol.~16, no.~12, p. 3005, 2007.

\bibitem{morin2013semi}
R.~Morin, S.~Bidon, A.~Basarab, and D.~Kouame, ``Semi-blind deconvolution for
  resolution enhancement in ultrasound imaging,'' in \emph{Image Processing
  (ICIP), 2013 20th IEEE International Conference on}, Sept 2013, pp.
  1413--1417.

\bibitem{zhao2014restoration}
N.~Zhao, A.~Basarab, D.~Kouame, and J.-Y. Tourneret, ``Restoration of
  ultrasound images using a hierarchical bayesian model with a generalized
  gaussian prior,'' in \emph{Image Processing (ICIP), 2014 IEEE International
  Conference on}, Oct 2014, pp. 4577--4581.

\bibitem{alessandrini2011restoration}
M.~Alessandrini, S.~Maggio, J.~Por{\'e}e, L.~De~Marchi, N.~Speciale,
  E.~Franceschini, O.~Bernard, and O.~Basset, ``A restoration framework for
  ultrasonic tissue characterization,'' \emph{Ultrasonics, Ferroelectrics, and
  Frequency Control, IEEE Transactions on}, vol.~58, no.~11, pp. 2344--2360,
  2011.

\bibitem{alessandrini2011statistical}
M.~Alessandrini, ``Statistical methods for analysis and processing of medical
  ultrasound: applications to segmentation and restoration,'' Ph.D.
  dissertation, University of Bologna, Bologna, Italy, 2011.

\bibitem{jensen1993deconvolution}
J.~A. Jensen, J.~Mathorne, T.~Gravesen, and B.~Stage, ``Deconvolution of in
  vivo ultrasound \uppercase{B}-mode images,'' \emph{Ultrasonic Imaging},
  vol.~15, no.~2, pp. 122--133, 1993.

\bibitem{ma2009deblurring}
J.~Ma and F.-X. Le~Dimet, ``Deblurring from highly incomplete measurements for
  remote sensing,'' \emph{Geoscience and Remote Sensing, IEEE Transactions on},
  vol.~47, no.~3, pp. 792--802, 2009.

\bibitem{xiao2011compounded}
L.~Xiao, J.~Shao, L.~Huang, and Z.~Wei, ``Compounded regularization and fast
  algorithm for compressive sensing deconvolution,'' in \emph{Image and
  Graphics (ICIG), 2011 Sixth International Conference on}.\hskip 1em plus
  0.5em minus 0.4em\relax IEEE, 2011, pp. 616--621.

\bibitem{zhao2010compressed}
M.~Zhao and V.~Saligrama, ``On compressed blind de-convolution of filtered
  sparse processes,'' in \emph{Acoustics Speech and Signal Processing (ICASSP),
  2010 IEEE International Conference on}.\hskip 1em plus 0.5em minus
  0.4em\relax IEEE, 2010, pp. 4038--4041.

\bibitem{amizic2013compressive}
B.~Amizic, L.~Spinoulas, R.~Molina, and A.~K. Katsaggelos, ``Compressive blind
  image deconvolution,'' \emph{Image Processing, IEEE Transactions on},
  vol.~22, no.~10, pp. 3994--4006, 2013.

\bibitem{spinoulas2012simultaneous}
L.~Spinoulas, B.~Amizic, M.~Vega, R.~Molina, and A.~K. Katsaggelos,
  ``Simultaneous bayesian compressive sensing and blind deconvolution,'' in
  \emph{Signal Processing Conference (EUSIPCO), 2012 Proceedings of the 20th
  European}.\hskip 1em plus 0.5em minus 0.4em\relax IEEE, 2012, pp. 1414--1418.

\bibitem{hegde2009compressive}
C.~Hegde and R.~G. Baraniuk, ``Compressive sensing of streams of pulses,'' in
  \emph{Communication, Control, and Computing, 2009. Allerton 2009. 47th Annual
  Allerton Conference on}.\hskip 1em plus 0.5em minus 0.4em\relax IEEE, 2009,
  pp. 44--51.

\bibitem{5677611}
C.~Hegde and R.~Baraniuk, ``Sampling and recovery of pulse streams,''
  \emph{Signal Processing, IEEE Transactions on}, vol.~59, no.~4, pp.
  1505--1517, April 2011.

\bibitem{boyd2011distributed}
\BIBentryALTinterwordspacing
S.~Boyd, N.~Parikh, E.~Chu, B.~Peleato, and J.~Eckstein, ``Distributed
  optimization and statistical learning via the alternating direction method of
  multipliers,'' \emph{Found. Trends Mach. Learn.}, vol.~3, no.~1, pp. 1--122,
  Jan. 2011. [Online]. Available: \url{http://dx.doi.org/10.1561/2200000016}
\BIBentrySTDinterwordspacing

\bibitem{zhao2015alternating}
X.~Zhao, C.~Chen, and M.~Ng, ``Alternating direction method of multipliers for
  nonlinear image restoration problems,'' \emph{Image Processing, IEEE
  Transactions on}, vol.~24, no.~1, pp. 33--43, Jan 2015.

\bibitem{jirik2008blind}
R.~Jirik and T.~Taxt, ``Two-dimensional blind bayesian deconvolution of medical
  ultrasound images,'' \emph{Ultrasonics, Ferroelectrics, and Frequency
  Control, IEEE Transactions on}, vol.~55, no.~10, pp. 2140--2153, 2008.

\bibitem{yu2012blind}
C.~Yu, C.~Zhang, and L.~Xie, ``A blind deconvolution approach to ultrasound
  imaging,'' \emph{Ultrasonics, Ferroelectrics, and Frequency Control, IEEE
  Transactions on}, vol.~59, no.~2, pp. 271--280, 2012.

\bibitem{7174535}
A.~Achim, A.~Basarab, G.~Tzagkarakis, P.~Tsakalides, and D.~Kouame,
  ``Reconstruction of ultrasound rf echoes modeled as stable random
  variables,'' \emph{Computational Imaging, IEEE Transactions on}, vol.~1,
  no.~2, pp. 86--95, June 2015.

\bibitem{michailovich2005novel}
O.~V. Michailovich and D.~Adam, ``A novel approach to the 2-d blind
  deconvolution problem in medical ultrasound,'' \emph{Medical Imaging, IEEE
  Transactions on}, vol.~24, no.~1, pp. 86--104, 2005.

\bibitem{pesquet2012parallel}
J.-C. Pesquet and N.~Pustelnik, ``A parallel inertial proximal optimization
  method,'' \emph{Pacific Journal of Optimization}, vol.~8, no.~2, pp.
  273--305, 2012.

\bibitem{pustelnik2011parallel}
N.~Pustelnik, C.~Chaux, and J.-C. Pesquet, ``Parallel proximal algorithm for
  image restoration using hybrid regularization,'' \emph{Image Processing, IEEE
  Transactions on}, vol.~20, no.~9, pp. 2450--2462, 2011.

\bibitem{pustelnik2012relaxing}
N.~Pustelnik, J.~Pesquet, and C.~Chaux, ``Relaxing tight frame condition in
  parallel proximal methods for signal restoration,'' \emph{Signal Processing,
  IEEE Transactions on}, vol.~60, no.~2, pp. 968--973, 2012.

\bibitem{combettes2011proximal}
P.~L. Combettes and J.-C. Pesquet, ``Proximal splitting methods in signal
  processing,'' in \emph{Fixed-point algorithms for inverse problems in science
  and engineering}.\hskip 1em plus 0.5em minus 0.4em\relax Springer, 2011, pp.
  185--212.

\bibitem{chouzenoux2014variable}
E.~Chouzenoux, J.-C. Pesquet, and A.~Repetti, ``Variable metric
  forward--backward algorithm for minimizing the sum of a differentiable
  function and a convex function,'' \emph{Journal of Optimization Theory and
  Applications}, vol. 162, no.~1, pp. 107--132, 2014.

\bibitem{he2002new}
B.~He, L.-Z. Liao, D.~Han, and H.~Yang, ``A new inexact alternating directions
  method for monotone variational inequalities,'' \emph{Mathematical
  Programming}, vol.~92, no.~1, pp. 103--118, 2002.

\bibitem{yang2011alternating}
J.~Yang and Y.~Zhang, ``Alternating direction algorithms for l1-problems in
  compressive sensing,'' \emph{SIAM journal on scientific computing}, vol.~33,
  no.~1, pp. 250--278, 2011.

\bibitem{deng2013group}
\BIBentryALTinterwordspacing
W.~Deng, W.~Yin, and Y.~Zhang, ``Group sparse optimization by alternating
  direction method,'' vol. 8858, 2013, pp. 88\,580R--88\,580R--15. [Online].
  Available: \url{http://dx.doi.org/10.1117/12.2024410}
\BIBentrySTDinterwordspacing

\bibitem{do2012fast}
T.~T. Do, L.~Gan, N.~H. Nguyen, and T.~D. Tran, ``Fast and efficient
  compressive sensing using structurally random matrices,'' \emph{Signal
  Processing, IEEE Transactions on}, vol.~60, no.~1, pp. 139--154, 2012.

\bibitem{ng2007wavelet}
J.~Ng, R.~Prager, N.~Kingsbury, G.~Treece, and A.~Gee, ``Wavelet restoration of
  medical pulse-echo ultrasound images in an em framework,'' \emph{Ultrasonics,
  Ferroelectrics, and Frequency Control, IEEE Transactions on}, vol.~54, no.~3,
  pp. 550--568, 2007.

\bibitem{jensen1991model}
J.~A. Jensen, ``A model for the propagation and scattering of ultrasound in
  tissue,'' \emph{Acoustical Society of America. Journal}, vol.~89, no.~1, pp.
  182--190, 1991.

\bibitem{wang2004image}
Z.~Wang, A.~C. Bovik, H.~R. Sheikh, and E.~P. Simoncelli, ``Image quality
  assessment: from error visibility to structural similarity,'' \emph{Image
  Processing, IEEE Transactions on}, vol.~13, no.~4, pp. 600--612, 2004.

\bibitem{combettes2005signal}
P.~L. Combettes and V.~R. Wajs, ``Signal recovery by proximal forward-backward
  splitting,'' \emph{Multiscale Modeling \& Simulation}, vol.~4, no.~4, pp.
  1168--1200, 2005.

\bibitem{raguet2013generalized}
H.~Raguet, J.~Fadili, and G.~Peyr{\'e}, ``A generalized forward-backward
  splitting,'' \emph{SIAM Journal on Imaging Sciences}, vol.~6, no.~3, pp.
  1199--1226, 2013.

\bibitem{lyshchik2005elastic}
A.~Lyshchik, T.~Higashi, R.~Asato, S.~Tanaka, J.~Ito, M.~Hiraoka, A.~Brill,
  T.~Saga, and K.~Togashi, ``Elastic moduli of thyroid tissues under
  compression,'' \emph{Ultrasonic imaging}, vol.~27, no.~2, pp. 101--110, 2005.

\end{thebibliography}
\end{document}